\documentclass[lettersize,journal]{IEEEtran}
\usepackage{amsmath,amsfonts,amssymb}
\usepackage{algorithmic}
\usepackage{algorithm}
\usepackage{array}
\usepackage{hyperref}
\usepackage{textcomp}
\usepackage{stfloats}
\usepackage{url}
\usepackage{verbatim}
\usepackage{graphicx}
\usepackage{subfigure}
\usepackage{cite}
\usepackage{xurl}
\usepackage{multirow}
\usepackage{tikz}

\begin{document}

\title{Privacy-Preserving Generation Fraud Detection for Distributed Photovoltaic Systems: A Solar Irradiance-Fused Federated Learning Framework}

\author{Xiaolu~Chen,
        Chenghao~Huang,
        Yanru~Zhang,~\IEEEmembership{Senior Member,~IEEE}
        and~Hao~Wang,~\IEEEmembership{Member,~IEEE}
\thanks{This work was supported in part by the Australian Research Council (ARC) Discovery Early Career Researcher Award (DECRA) under Grant DE230100046. (Corresponding author: Hao Wang, Yanru Zhang)}
\thanks{X. Chen is with the School of Computer Science and Technology, University of Electronic Science and Technology of China, Chengdu, China and the Department of Data Science and AI, Faculty of IT, Monash University, Melbourne, VIC 3800, Australia (e-mail: jzzcqebd@gmail.com; xiaolu.chen@monash.edu).}
\thanks{C. Huang and H. Wang are with the Department of Data Science and AI, Faculty of IT and Monash Energy Institute, Monash University, Melbourne, VIC 3800, Australia (e-mails: \{chenghao.huang, hao.wang2\}@monash.edu).}
\thanks{Y. Zhang is with the School of Computer Science and Technology, University of Electronic Science and Technology of China, Chengdu, and Shenzhen Institute for Advanced Study of UESTC, Shenzhen, China (e-mail: yanruzhang@uestc.edu.cn).}
}

\maketitle

\begin{abstract}
The wide adoption of residential photovoltaic (PV) systems introduces new challenges for generation fraud detection (FD). Unlike traditional electricity theft detection, which focuses on electricity consumption-side behavior, PV generation fraud detection (PVG-FD) is complicated by the inherent intermittency and uncertainty of PV generation.
The distributed nature of PV systems poses further challenges for centralized PVG-FD approaches due to scalability and privacy concerns. This paper develops a privacy-preserving distributed PVG-FD framework based on federated learning (FL). In this framework, a utility company manages multiple household communities, where each of which is equipped with a local detector. 
The framework integrates a novel detection model architecture with privacy-preserving global collaboration. 
Each community's local model fuses PV generation and weather data via a co-attention mechanism to detect discrepancies critical for PVG-FD. 
The FL framework enables cross-community collaboration by aggregating model parameters and prototypes, leveraging global knowledge sharing with local refinement while preserving privacy. It also uses prototype alignment to address class imbalance by enhancing fraud sample representation.
Extensive experiments on a real-world residential PV dataset validate the effectiveness of the developed method and demonstrate that it outperforms state-of-the-art FL methods across various scenarios. The results also show its scalability across varying community sizes and strong robustness to class imbalance.
\end{abstract}


\section{Introduction}

\subsection{Background and Motivation}
Renewable energy sources, particularly solar photovoltaics (PV), have experienced rapid growth in recent years~\cite{IEA_news1}. Distributed PV systems are expected to account for nearly half of the total solar PV deployment, exceeding the total deployment of onshore wind power over the same period~\cite{IEA_news2}. As distributed PV continues to expand globally, its monitoring and management have become increasingly important.
Advanced metering infrastructure (AMI) serves as a fundamental component of smart grids, enabling efficient energy monitoring and management. However, inherent security vulnerabilities expose modern power systems to significant risks~\cite{Xu2020}. Cyber-attackers can exploit the limited tamper resistance of these devices to manipulate meter readings, thereby committing electricity theft~\cite{9686052}.
Beyond conventional electricity theft, the increasing integration of renewable energy into the power system has introduced novel challenges. Prosumers with distributed renewable generation systems make profits by selling their electricity to utility companies. However, some malicious actors manipulate meter data to overclaim their revenue, exceeding their actual generation~\cite{8998142}.
According to research analysis in~\cite{8355252}, strategically overreporting PV generation to evade detection can reduce the time required to recover the cost of installing residential rooftop PV systems by approximately 80\%.
This emerging form of generation fraud not only causes financial losses for utility companies but also disrupts the balance between supply and demand in the energy system~\cite{9696293}, affecting power system reliability and stability.
Addressing these challenges requires specialized detection methods tailored to the unique characteristics of renewable energy systems.

In recent years, efforts have focused on detecting PV generation fraud. Some studies rely solely on meter data that capture PV generation measurements~\cite{10322383,9964082,CHEN2025115461,9712854}, without considering additional factors such as weather conditions, which are crucial to PV generation efficiency. Others incorporate weather data as auxiliary information to implicitly capture generation behavior under physical constraints, allowing detectors to identify overreporting~\cite{8998142,9528289,8355252,TANG2023109212,11004263}. However, existing methods that combine multiple data sources may struggle to capture the nonlinear and dynamic relationship between PV generation and weather conditions.

Meanwhile, these multi-source data approaches typically adopt centralized model training paradigms that require aggregating large-scale user meter data from AMI. 
This centralized approach suffers from several limitations, including high communication and computation costs, as well as risks of data privacy breaches~\cite{9696293}. Federated learning (FL), a distributed learning paradigm, has emerged as a promising solution to address these issues by enabling collaborative model training without directly sharing local data~\cite{pmlr-v54-mcmahan17a,10246323}. 
However, employing FL for detecting PV generation fraud presents several challenges.
\begin{itemize}
    \item \textbf{Data Modality Gap}: 
    Although both PV generation data and weather data are represented as time series, they differ fundamentally in physical meaning, measurement granularity, and noise characteristics. PV generation data are device-level outputs influenced by installation-specific factors and local weather, whereas weather data reflect regional environmental conditions that may not perfectly match the conditions at an individual installation. These differences introduce a semantic gap between the two modalities, making it challenging for conventional multi-source data approaches to jointly model them. Therefore, the nonlinear and dynamic relationships between PV generation and weather data are not fully captured, impairing detection performance.
    \item \textbf{Class Imbalance}: Fraudulent generation cases constitute only a small fraction of PV generation fraud detection (PVG-FD) datasets, with most cases reflecting lawful, normal behaviors. This imbalance causes model training to focus more on normal cases, making it difficult to learn effective features for fraudulent cases and thereby reducing detection performance.
    In distributed, multi-community settings, each community holds only a subset of the overall data, so the absolute number of fraudulent cases per community is even smaller, further hindering the learning of discriminative fraud-related features.
\end{itemize}

\subsection{Literature Review}
To situate our work within existing studies, we review related works on PVG-FD and related FL methods.

\subsubsection{PV Generation Fraud Detection}
PV generation fraud can be inferred from household-level time-series meter data. A variety of data-driven methods have therefore been proposed to identify manipulated generation patterns from meter data.

One of the key early studies~\cite{8998142} explicitly formulated PV generation attack functions, providing a definition of how fraudulent generation can be produced that anchors subsequent fraud-detection research. Traditional feature-engineering and machine-learning approaches have also been explored for PVG-FD, including models based on engineered PV features~\cite{10322383}, clustering-driven detection~\cite{9632322}, tree-based models~\cite{10.1145/3447555.3464852,9528289}, and a quantum-inspired classifier~\cite{10949078}. These methods offer early data-driven solutions but remain constrained by handcrafted features and limited temporal modeling capacity.

Research then progressed toward more expressive deep learning architectures and richer feature representations, including the use of contextual information, such as weather conditions, to help distinguish abnormal PV behaviors from natural environmental variability. Neural network detectors, such as feedforward models in~\cite{10199139}, were complemented by recurrent approaches that leveraged gated recurrent unit (GRU) and bidirectional long short-term memory (Bi-LSTM) architectures with robustness enhancement~\cite{9964082}. More convolutional architectures have also been explored, ranging from a convolutional neural network (CNN)–based detector~\cite{TANG2023109212}, through a graph convolutional network (GCN)–based model~\cite{LIU2025110551}, to a deeper residual network (ResNet) architecture~\cite{BHUSAL2022107345}. Hybrid architectures that combine convolutional networks with recurrent or attention-based modules have also been investigated, including a CNN–Bi-LSTM model~\cite{CHEN2025115461} and a multiscale CNN–LSTM–Transformer hybrid designed to capture long-range temporal dependencies~\cite{11004263}.
Despite these advances, most existing methods are built on centralized architectures with inherent privacy concerns and do not explicitly model the dependencies between PV generation and associated irradiance measurements, leaving substantial room for improvement.

\subsubsection{Federated Learning for Electricity Theft Detection}
FL has emerged as an important paradigm for electricity theft detection, as it allows multiple data holders to collaboratively train detection models without sharing raw data. Motivated by these benefits, researchers have developed a variety of FL-based detection methods, ranging from foundational frameworks to deep learning–enhanced models.

Early FL studies primarily focused on developing privacy-preserving collaborative modeling for consumption-based electricity theft detection. Wang et al.~\cite{10246323} and Han et al.~\cite{10512760} applied FL to this task by introducing a federated XGBoost classifier and a federated CNN-based detector, respectively, trained across multiple independent data holders. Guo et al.~\cite{10807710} developed a sequential decentralized FL framework with progressive channel pruning to reduce communication and inference costs, while Li et al.~\cite{li2024fuse} presented a resource-efficient FL architecture integrating a three-tier U-shape split-learning scheme.
Subsequent studies introduced more expressive deep learning architectures into FL for consumption-based electricity theft detection. Several works adopted CNN-based hybrid models, including the CNN–GRU detector~\cite{10000881}, the CNN–BiGRU–attention model~\cite{10191166}, and the CNN–LSTM–based approach~\cite{WEN2025124789}. Wen et al.~\cite{9531953} introduced a temporal convolutional network (TCN)–based federated detector to capture multiscale consumption patterns. Building on recurrent modeling capacity, Wang et al.~\cite{WANG2024109848} adopted a stacked GRU ensemble within an FL framework to increase modeling flexibility for complex consumption dynamics.
Existing FL-based electricity theft detection studies target anomalies arising from load-data manipulation. Their problem formulation and feature design are specific to residential consumption behaviors and therefore do not transfer to PV generation fraud scenarios.

\subsubsection{Addressing Data Imbalance in PV Generation Fraud Detection and Federated Learning}

Class imbalance is a critical challenge in PVG-FD, as fraudulent behaviors are typically rare compared to normal generation cases.
The same minority-class scarcity also arises in federated settings, where data remain decentralized across clients. We therefore review imbalance-handling methods in both centralized PVG-FD and FL.

PVG-FD studies have adopted a variety of strategies to mitigate the scarcity of minority-class fraud samples. Data augmentation and synthetic sample generation have been used, including approaches that create realistic temporal fraud patterns~\cite{9712854} or leverage generative models, such as balanced temporal GANs~\cite{LIU2025110551}. Other works address imbalance through modeling choices or loss design, such as one-class detectors trained solely on benign data~\cite{9528289}, loss reweighting for minority-class emphasis~\cite{10199139}, and gradient-level balancing techniques~\cite{CHEN2025115461}.
These techniques address class scarcity in centralized PV fraud detection, providing useful context for understanding how imbalance is handled before considering federated settings.
FL research has also explored techniques for improving minority-class performance when data are locally imbalanced across clients. In the context of electricity theft detection, some studies incorporate imbalance-aware losses into the local training objective, such as focal loss~\cite{10246323} or multi-level weighted schemes that amplify the contribution of under-represented samples during federated optimization~\cite{WEN2025124789}. Beyond application-level detectors, a series of FL algorithms also address class imbalance through representation learning and aggregation design, including global label-centroid sharing with contrastive objectives~\cite{11150383}, client-side self-balancing mechanisms~\cite{9825928}, and prototype-based approaches~\cite{Dai_Chen_Li_Heinecke_Sun_Xu_2023}. Together, these works illustrate a broader landscape of imbalance-aware strategies in federated environments.
Despite these efforts, existing imbalance-aware methods in both centralized PVG-FD and federated settings address different problem structures and data assumptions, and thus do not directly apply to PVG-FD in federated environments.

\subsection{Main Work and Contributions}

\begin{table*}[h!tbp]
\centering
\caption{Summary of our work against existing related studies.}
\label{tab:intro_summary}
\renewcommand{\arraystretch}{1.08}
\resizebox{\textwidth}{!}{
\begin{tabular}{l c c c}
\hline
\textbf{Study category}           & \textbf{Privacy-preserving distributed learning} & \textbf{PVG-FD-specific design} & \textbf{Imbalance-aware design} \\ \hline
Centralized PVG-FD methods        & $\times$                                         & \checkmark                      & Partial                              \\
FL-based electricity theft detection methods              & \checkmark                                       & $\times$                        & Partial                              \\
Imbalance-aware PVG-FD/FL methods & Partial                                          & Partial                         & \checkmark                           \\
This work                         & \checkmark                                       & \checkmark                      & \checkmark                          \\ \hline
\end{tabular}
}
\end{table*}

As shown in Table~\ref{tab:intro_summary}, several key research gaps remain in existing studies on PVG-FD. First, existing PVG-FD methods are mainly developed under centralized settings, which limits their applicability in privacy-sensitive multi-community scenarios. Second, existing FL-based electricity theft detection methods are not specifically designed for PVG-FD, due to the distinct data characteristics and anomaly patterns of photovoltaic generation fraud. Third, although imbalance-aware methods have been studied in both PVG-FD and FL, effective solutions tailored to scarce fraud samples in federated PVG-FD remain underexplored.

To address these gaps, we propose a federated framework for privacy-preserving PVG-FD. The framework consists of locally trained residential-level detection models and a global aggregation mechanism that enables collaborative learning across communities without sharing raw data. Within this framework, we further develop a multi-source data fusion design to jointly model PV generation and weather information, and incorporate a cross-community knowledge-sharing mechanism to mitigate the impact of scarce fraud samples in federated PVG-FD.

The contributions of this work are as follows.
\begin{itemize}
    \item To enable privacy-preserving collaborative training for PVG-FD across communities, we propose a distributed method based on the FL paradigm with a focus on addressing class-imbalance characteristics inherent in the PVG-FD problem. Experimental results demonstrate superior performance over several baselines.
    \item To improve detection performance under varying environmental conditions, we design a residential-level detection model based on multi-source data fusion. The model fuses PV generation and multivariate irradiance data through a co-attention mechanism to capture complex cross-modal dependencies.
    \item To mitigate the impact of scarce generation fraud samples on local model training, we introduce a prototype-based regularization mechanism into the FL framework. This design enhances fraud-related representation learning and promotes privacy-preserving cross-community knowledge sharing, thereby improving PVG-FD performance.
\end{itemize}

The remainder of this paper is organized as follows. Section \ref{Sec:PS} presents the problem statement of PVG-FD. Section \ref{Sec:M} details the design of the privacy-perserving distributed PVG-FD framework using FL. Section \ref{Sec:E} presents the experimental datasets, performance metrics, benchmarks, and results. Section \ref{Sec:C} concludes this paper.

\section{Problem Statement}\label{Sec:PS}

\begin{figure}[t]
\centering
\includegraphics[width=\columnwidth]{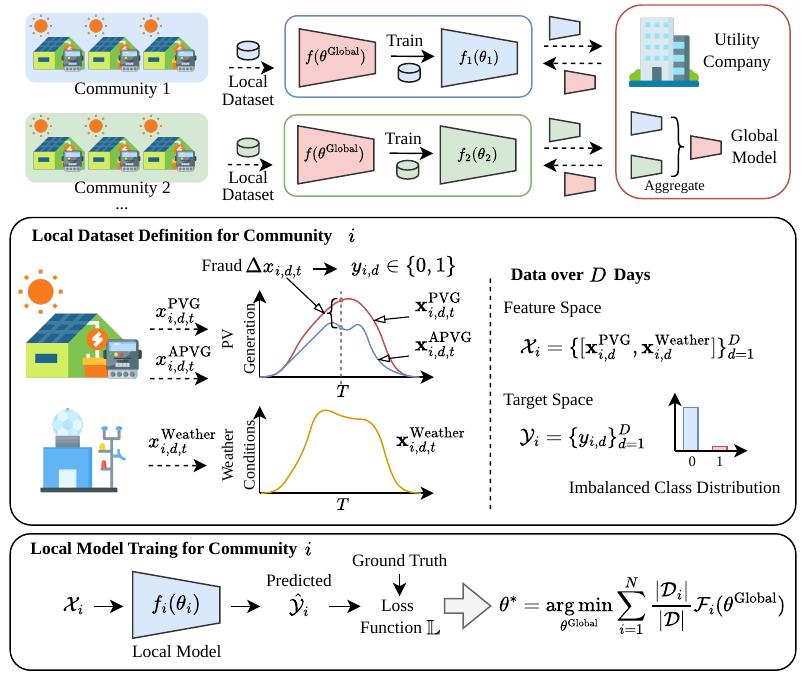}
\caption{The FL paradigm for distributed PVG-FD.}
\label{fig:ps_2}
\end{figure}

We consider a utility company serving $N$ communities, where the $i$-th community owns a local private dataset $\mathcal{D}_i$. 
Each community has multiple prosumers and runs a residential-level PVG-FD model, which is designed to detect instances of PV generation fraud by any prosumer within the community on a given day.
A distributed PVG-FD framework based on the FL paradigm enables residential-level models to enhance performance without sharing raw data, as shown in Fig.~\ref{fig:ps_2}.

In the $i$-th community, for each day $d$ in a total of $D$ days, a daily horizon is divided into $T$ discrete time slots. We denote the actual generation of a solar prosumer at time slot $t$ as $x_{i,d,t}^{\text{APVG}}$ and the generation reported to the utility company via a smart meter as $x^{\text{PVG}}_{i,d,t}$. Under normal conditions, these two values coincide, i.e., $x_{i,d,t}^{\text{PVG}}=x^{\text{APVG}}_{i,d,t}$. However, in cases of PV generation fraud, the prosumer may inflate the reported generation to gain illicit profits, which can be formulated as
\begin{align}
    x^{\text{PVG}}_{i,d,t} = x_{i,d,t}^{\text{APVG}} + \Delta x_{i,d,t},
\end{align}
where $\Delta x_{i,d,t}>0$ denotes the fraudulently reported output.

PV generation fraud typically occurs continuously or periodically over extended periods rather than in a single time slot. To reduce misjudgments from external factors and better identify persistent fraud patterns, utility companies should analyze data at a daily timescale. 
Accordingly, we define a daily reported-generation time series as $\mathbf{x}^{\text{PVG}}_{i,d} = \{x^{\text{PVG}}_{i,d,t}\}^T_{t=1}$. 
To indicate whether fraud occurs on a given day $d$ in the $i$-th community, we introduce a binary variable $y_{i,d}$ defined by
\begin{align}
    y_{i,d}=\begin{cases} 
    1, & \text{if fraud occurs on day}\ \ d, \\
    0 ,& \text{otherwise.}
    \end{cases}
\end{align}
Furthermore, as the characteristics of individual PV panels may remain unknown, it is beneficial to incorporate weather conditions to enhance detection performance. 
Thus, we include a daily weather time series $\mathbf{x}^{\text{Weather}}_{i,d} = \{x^{\text{Weather}}_{i,d,t}\}^T_{t=1}$ as auxiliary information. 

Accordingly, the feature space $\mathcal{X}_{i}$ consists of both reported generation and weather conditions across all $D$ recorded days of the $i$-th community, represented as
\begin{align}
    \mathcal{X}_{i} = \{[\mathbf{x}^{\text{PVG}}_{i,d},\mathbf{x}^{\text{Weather}}_{i,d}]\}_{d=1}^D \in \mathbb{R}^{D\times 2\times T}.
\end{align}
The target space $\mathcal{Y}_{i}$, on the other hand, contains the corresponding ground-truth indicators of whether fraud occurs on each day:
\begin{align}
    \mathcal{Y}_{i} = \{y_{i,d}\}_{d=1}^D \in \mathbb{R}^{D}.
\end{align}

Then, we denote the collection of all local datasets as $\mathcal{D} = \{\mathcal{D}_i\}_{i=1}^{N}$.
For $\mathcal{D}_i$, the samples and labels are represented as $\mathcal{X}_i$ and $\mathcal{Y}_i$, respectively. Thus, the empirical risk for the $i$-th community can be formulated as
\begin{align}
\mathcal{F}_i(\theta_i):=\mathbb{E}_{(\mathcal{X}_i,\mathcal{Y}_i) \sim \mathcal{D}_i}\Big[\mathbb{L}\big[f_i(\theta_i;\mathcal{X}_i),\mathcal{Y}_i\big]\Big],
\end{align}
where $f_i(\theta_i;\cdot)$ denotes the model with parameters $\theta_i$, and $\mathbb{L}$ denotes the loss function associated with the PVG-FD task, measuring the discrepancy between the model's predictions $\hat{\mathcal{Y}}$ and the ground truth $\mathcal{Y}$. Note that, in our work, model heterogeneity is not considered. Thus, the model architectures of each community are the same:
\begin{align}
    f_1(\cdot) = f_2(\cdot) = \cdots = f_N(\cdot).
\end{align}

In the FL framework, all communities upload their local models to the utility company. Then, the utility company aggregates these models into a shared global model $f(\theta^{\text{Global}})$ that maintains the architecture $f(\cdot)$ consistent with all the uploaded local models, and a set of parameters $\theta^{\text{Global}}$. This global model generates predictions $\hat{\mathcal{Y}}_i = f(\theta^{\text{Global}}; \mathcal{X}_i)$, which should approximate the true output $\mathcal{Y}_i$. 

Let $|\mathcal{D}_i|$ represent the size of the $i$-th dataset, and $|\mathcal{D}|$ denote the total size of all the datasets. The overall training objective, in line with methods such as FedAvg, is to find optimal global model parameters $\theta^{*}$ by minimizing the weighted sum of the empirical risks across all local datasets:
\begin{align}
    \theta^* = \mathop{\arg \min}\limits_{\theta^{\text{Global}}} \sum_{i=1}^{N} \frac{|\mathcal{D}_i|}{|\mathcal{D}|} \mathcal{F}_i(\theta^\text{Global}).
\end{align}
In this way, the global model is improved using distributed datasets, ensuring that the specific characteristics of each community's data are incorporated without sharing raw information.
This problem formulation poses two challenges. Firstly, directly fusing the input features is difficult due to their distinct data modalities.
Secondly, the class distribution in the target space, where normal samples significantly outnumber generation fraud samples, is highly imbalanced and further complicates model learning.
To address these challenges, we incorporate multi-source data fusion and representation learning in the FL framework, which are detailed in Section~\ref{Sec:M}.

\section{Proposed Privacy-Preserving Distributed PVG-FD Framework}\label{Sec:M}
Detecting PV generation fraud in a distributed setting presents two major challenges: (1) how to effectively model the complex dependencies between PV generation and environmental conditions using decentralized data, and (2) how to collaboratively train detection models across communities while addressing class imbalance. 
To tackle these issues, we develop a two-part methodology: a residential-level detection model based on multi-source data fusion, and an FL framework that supports collaborative training without sharing raw data. The former enhances local detection capability by learning informative representations from PV generation and solar irradiance data. The latter facilitates global knowledge sharing while mitigating class imbalance through prototype-guided regularization. In the following subsections, we detail the design of the local detection model and the federated training framework.

\subsection{Residential-level PV Generation Fraud Detection Model}

\begin{figure*}[t]
\centering
\includegraphics[width=0.95\textwidth]{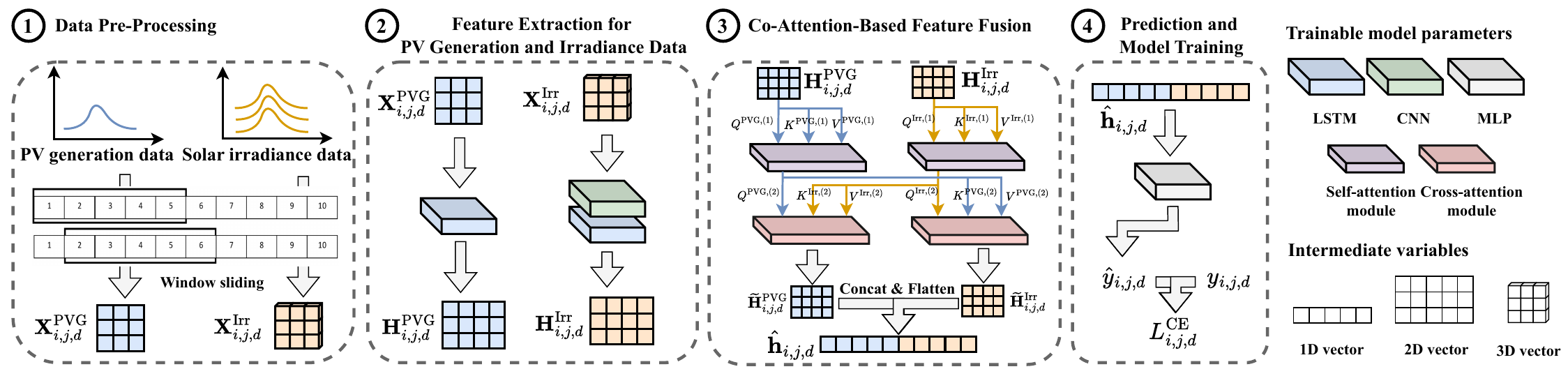}
\caption{The architecture of the $i$-th community's PVG-FD model carrying out detection on the $d$-th day of the $j$-th prosumer.} 
\label{fig:local_model}
\end{figure*}

The developed PVG-FD model, as shown in Fig.~\ref{fig:local_model}, is designed as the local detection model in our FL framework, enabling each community to perform fraud detection using its own data without sharing raw data.
The model consists of four key modules: data pre-processing, feature extraction for PV generation and irradiance data, co-attention-based feature fusion, and prediction and model training. By sequentially processing the inputs through these modules, the model captures complementary patterns from two data modalities and updates its parameters via gradient descent based on the loss function.

\subsubsection{Data Pre-Processing}

At the local end, each community performs PVG-FD using recent time-series data from smart meter readings. Due to privacy constraints, detailed PV system specifications are not accessible.
To compensate for this lack of system-level information, external weather data, including diffuse horizontal irradiance (DHI), global horizontal irradiance (GHI), and direct normal irradiance (DNI), are incorporated, as these variables closely relate to PV generation.

To capture local temporal patterns and enhance feature extraction, a sliding window approach is utilized to segment each prosumer's daily PV generation sequence into overlapping sub-sequences (windows) of a fixed length $T^{\text{Sub}}$, using a stride $S < T^{\text{Sub}}$ to control the overlap. 
On the $d$-th day, the input data for the $j$-th prosumer in the $k$-th window are defined as
\begin{align}
    \mathbf{x}_{i,j,d,k}^{\text{PVG}}=\{x^{\text{PVG}}_{i,j,d,t}\}^{(k-1)S+T^\text{Sub}}_{t=1+(k-1)S}, k \in \{1, \cdots, L^\text{Window}\},
\end{align}
where $L^\text{Window}=\frac{T-T^{\text{Sub}}}{S}+1$ is the total number of windows.
The segmented daily PV generation sequence is expressed as
\begin{align}
\mathbf{X}_{i,j,d}^{\text{PVG}}=(\mathbf{x}_{i,j,d,1}^{\text{PVG}};\dots;\mathbf{x}_{i,j,d,L^{\text{Window}}}^{\text{PVG}})^\top \in \mathbb{R}^{T^{\text{Sub}}\times L^{\text{Window}}},\\
i \in \{1,...,N\}, \quad j\in\{1,...,M_i\}, \quad d \in \{1,...,D\}. \notag
\end{align}

Similarly, the segmented irradiance data is formulated as
\begin{align}
\mathbf{X}_{i,j,d}^{\text{Irr}}=\begin{bmatrix}
(\mathbf{x}_{i,j,d,1}^{\text{DHI}};\dots;\mathbf{x}_{i,j,d,L^{\text{Window}}}^{\text{DHI}})^\top\\(\mathbf{x}_{i,j,d,1}^{\text{DNI}};\dots;\mathbf{x}_{i,j,d,L^{\text{Window}}}^{\text{DNI}})^\top\\(\mathbf{x}_{i,j,d,1}^{\text{GHI}};\dots;\mathbf{x}_{i,j,d,L^{\text{Window}}}^{\text{GHI}})^\top
\end{bmatrix}
 \in \mathbb{R}^{T^\text{Sub} \times L^{\text{Window}} \times 3}, \\
i \in \{1,...,N\}, \quad j\in\{1,...,M_i\}, \quad d \in \{1,...,D\}, \notag
\end{align}
where $\mathbf{x}_{i,j,d,k}^{\text{DHI}}\in\mathbb{R}^{T^{\text{Sub}}}$, $\mathbf{x}_{i,j,d,k}^{\text{DNI}}\in\mathbb{R}^{T^{\text{Sub}}}$, and $\mathbf{x}_{i,j,d,k}^{\text{GHI}}\in\mathbb{R}^{T^{\text{Sub}}}$ represent the $k$-th window of the DHI, DNI, and GHI sequences, respectively.

After segmentation, multivariate irradiance data are treated as image-like structures to exploit inter-channel correlations among DHI, GHI, and DNI using convolution.

\subsubsection{Feature Extraction for PV Generation and Irradiance Data}

To process the two different sources of data, we utilize two distinct feature extraction pipelines. For the $i$-th community, the solar PV generation data, represented as a univariate time series, are analyzed by a LSTM model. The localized temporal feature $\mathbf{H}_{i,j,d}^{\text{PVG}}$ is captured as
\begin{align}
    \mathbf{H}_{i,j,d}^{\text{PVG}} = \text{LSTM}_i (\mathbf{X}_{i,j,d}^{\text{PVG}}) \in \mathbb{R}^{{T^{\text{Sub}}}\times d_{\text{LSTM}}},
    \label{Eq:lstm}
\end{align}
where $d_{\text{LSTM}}$ is the hidden dimension of the LSTM.

For the irradiance data, represented as a multivariate time series with three features, i.e., DHI, GHI, and DNI, we employ a CNN-LSTM architecture.
For each day $d$, a 2D convolution operation is applied to extract spatial correlations among the three irradiance metrics. Then, the convolutional features are flattened and input into the following LSTM layers. Finally, one embedding feature per time step is obtained. 
Formally, we denote the CNN-LSTM as $f_i^{\text{Irr}}$, and the irradiance feature $\mathbf{H}^{\text{Irr}}_{i,j,d}$ is obtained as follows:
\begin{align}
    \mathbf{H}^{\text{Irr}}_{i,j,d} = \text{CNN-LSTM}_i(\mathbf{X}^{\text{Irr}}_{i,j,d}) \in \mathbb{R}^{{T^{\text{Sub}}}\times d_{\text{CNN-LSTM}}},
    \label{Eq:cnn-lstm}
\end{align}
where $d_{\text{CNN-LSTM}}$ is the hidden dimension of the CNN-LSTM.

\subsubsection{Co-Attention-Based Feature Fusion}
To fuse the extracted PV generation and irradiance features, we employ a co-attention mechanism.
The co-attention mechanism, which consists of self-attention~\cite{10.5555/3295222.3295349} and cross-attention~\cite{Lee_2018_ECCV} modules, 
fuses PV generation and irradiance feature spaces to capture inconsistencies indicative of PV generation fraud.

Initially, via self-attention, the features are linearly transformed into query, key, and value matrices for attention computation over the temporal dimension:
\begin{align}
    Q^{\text{PVG},(1)}=\mathbf{H}_{i,j,d}^{\text{PVG}} U_Q^{(1)}, \quad
    Q^{\text{Irr},(1)}=\mathbf{H}_{i,j,d}^{\text{Irr}} W_Q^{(1)}, \\ K^{\text{PVG},(1)}=\mathbf{H}_{i,j,d}^{\text{PVG}} U_K^{(1)}, \quad
     K^{\text{Irr},(1)}=\mathbf{H}_{i,j,d}^{\text{Irr}} W_K^{(1)}, \\ V^{\text{PVG},(1)}=\mathbf{H}_{i,j,d}^{\text{PVG}} U_V^{(1)}, \quad
     V^{\text{Irr},(1)}=\mathbf{H}_{i,j,d}^{\text{Irr}} W_V^{(1)},
\end{align}
where $U_Q^{(1)}, U_K^{(1)}, U_V^{(1)} \in \mathbb{R}^{d_{\text{LSTM}} \times d_{\text{SA}}}$ and $W_Q^{(1)},W_K^{(1)},W_V^{(1)} \in \mathbb{R}^{d_{\text{CNN-LSTM}} \times d_{\text{SA}}}$ are learnable parameters associated with the PV generation and irradiance features, respectively.

Next, two attention weight matrices are computed separately for the PV generation and irradiance features. The scaled dot product between the corresponding query and key matrices is calculated, and a softmax operation is applied to normalize the weights:
\begin{align}
    \mathbf{A}_{i,j,d}^{\text{G}} = \text{Softmax}\left(\frac{Q^{\text{PVG},(1)} (K^{\text{PVG},(1)})^\top}{\sqrt{d_{\text{SA}}}}\right) \in \mathbb{R}^{{T^{\text{Sub}}}\times {T^{\text{Sub}}}},\\
    \mathbf{A}_{i,j,d}^{\text{I}} = \text{Softmax}\left(\frac{Q^{\text{Irr},(1)} (K^{\text{Irr},(1)})^\top}{\sqrt{d_{\text{SA}}}}\right) \in \mathbb{R}^{{T^{\text{Sub}}}\times {T^{\text{Sub}}}}.
\end{align}
Finally, the self-attention-refined features $\widetilde{\mathbf{H}}_{i,j,d}^{\text{PVG},(1)}$ and $\widetilde{\mathbf{H}}_{i,j,d}^{\text{Irr},(1)}$ are obtained by performing a weighted summation of the value matrices, where the weights are determined by the respective attention matrices:
\begin{align}
    \widetilde{\mathbf{H}}_{i,j,d}^{\text{PVG},(1)} = \mathbf{A}_{i,j,d}^{\text{G}} V^{\text{PVG},(1)} \in \mathbb{R}^{{T^{\text{Sub}}} \times d_{\text{SA}}},\\
    \widetilde{\mathbf{H}}_{i,j,d}^{\text{Irr},(1)} =  \mathbf{A}_{i,j,d}^{\text{I}} V^{\text{Irr},(1)} \in \mathbb{R}^{{T^{\text{Sub}}}\times d_{\text{SA}}}.
\end{align}

The cross-attention module performs bi-directional cross-attention (i.e., co-attention), where PV generation attends to irradiance data and irradiance data attends to PV generation.

Similarly, to enable the features refined by the self-attention module to interact, they are linearly transformed into query, key, and value matrices for the cross-attention operation:
\begin{align}
    Q^{\text{PVG},(2)}=\widetilde{\mathbf{H}}_{i,j,d}^{\text{PVG},(1)} U_Q^{(2)}, \quad
    Q^{\text{Irr},(2)}=\widetilde{\mathbf{H}}_{i,j,d}^{\text{Irr},(1)} W_Q^{(2)}, \\ K^{\text{PVG},(2)}=\widetilde{\mathbf{H}}_{i,j,d}^{\text{PVG},(1)} U_K^{(2)}, \quad
     K^{\text{Irr},(2)}=\widetilde{\mathbf{H}}_{i,j,d}^{\text{Irr},(1)} W_K^{(2)}, \\ V^{\text{PVG},(2)}=\widetilde{\mathbf{H}}_{i,j,d}^{\text{PVG},(1)} U_V^{(2)}, \quad
     V^{\text{Irr},(2)}=\widetilde{\mathbf{H}}_{i,j,d}^{\text{Irr},(1)} W_V^{(2)},
\end{align}
where $U_Q^{(2)}, U_K^{(2)}, U_V^{(2)} \in \mathbb{R}^{d_{\text{SA}} \times d_{\text{CA}}}$ and $W_Q^{(2)},W_K^{(2)},W_V^{(2)} \in \mathbb{R}^{d_{\text{SA}} \times d_{\text{CA}}}$ are learnable parameters.

The cross-attention assigns weights to bidirectional dependencies: (1) PV generation features attending to irradiance features and (2) irradiance features attending to PV generation features.
The cross-attention weight matrices $\mathbf{A}_{i,j,d}^{\text{G-I}}$ and $\mathbf{A}_{i,j,d}^{\text{I-G}}$ are computed as:
\begin{align}
    \mathbf{A}_{i,j,d}^{\text{G-I}} = \text{Softmax}\left(\frac{Q^{\text{PVG},(2)} (K^{\text{Irr},(2)})^\top}{\sqrt{d_{\text{CA}}}}\right) \in \mathbb{R}^{{T^{\text{Sub}}}\times {T^{\text{Sub}}}},\\
    \mathbf{A}_{i,j,d}^{\text{I-G}} = \text{Softmax}\left(\frac{Q^{\text{Irr},(2)} (K^{\text{PVG},(2)})^\top}{\sqrt{d_{\text{CA}}}}\right) \in \mathbb{R}^{{T^{\text{Sub}}}\times {T^{\text{Sub}}}}.
\end{align}
The cross-attention outputs $\widetilde{\mathbf{H}}_{i,j,d}^{\text{PVG},(2)}$ and $\widetilde{\mathbf{H}}_{i,j,d}^{\text{Irr},(2)}$ are obtained by performing a weighted summation of the value matrices:
\begin{align}
    \widetilde{\mathbf{H}}_{i,j,d}^{\text{PVG},(2)} = \mathbf{A}_{i,j,d}^{\text{G-I}} V^{\text{Irr},(2)} \in \mathbb{R}^{{T^{\text{Sub}}} \times d_{\text{CA}}},\\
    \widetilde{\mathbf{H}}_{i,j,d}^{\text{Irr},(2)} =  \mathbf{A}_{i,j,d}^{\text{I-G}} V^{\text{PVG},(2)} \in \mathbb{R}^{{T^{\text{Sub}}} \times d_{\text{CA}}}.
\end{align}

For clarity, both the self-attention and cross-attention equations are presented in a single-head form; however, the actual implementation adopts multi-head attention with the number of heads $h$, following~\cite{10.5555/3295222.3295349}.
We denote the cross-attention outputs as 
$\widetilde{\mathbf{H}}_{i,j,d}^{\text{PVG}} := \widetilde{\mathbf{H}}_{i,j,d}^{\text{PVG},(2)}$ 
and 
$\widetilde{\mathbf{H}}_{i,j,d}^{\text{Irr}} := \widetilde{\mathbf{H}}_{i,j,d}^{\text{Irr},(2)}$.
The fused features $\widetilde{\mathbf{H}}_{i,j,d}^{\text{PVG}}$ and $ \widetilde{\mathbf{H}}_{i,j,d}^{\text{Irr}}$ are concatenated to produce a feature matrix and then flattened into a one-dimensional vector $\hat{\mathbf{h}}_{i,j,d}$:
\begin{align}
    \hat{\mathbf{H}}_{i,j,d} = \text{Concat}\left(\widetilde{\mathbf{H}}_{i,j,d}^{\text{PVG}}, \widetilde{\mathbf{H}}_{i,j,d}^{\text{Irr}}\right) \in \mathbb{R}^{{T^{\text{Sub}}} \times 2d_{\text{CA}}}, \\
    \hat{\mathbf{h}}_{i,j,d} = \text{Flatten}\left(\hat{\mathbf{H}}_{i,j,d}\right)
\in \mathbb{R}^{D^{p}}, \  D^{p} = T^{\text{Sub}} \cdot 2d_{\text{CA}}.
    \label{Eq:flatten}
\end{align}

\subsubsection{Prediction and Model Training}

The feature $\hat{\mathbf{h}}_{i,j,d}$ is utilized to predict the final judgment on the PV generation fraud behavior of the $j$-th prosumer in the $i$-th community at day $d$ by a multi-layer perceptron (MLP):
\begin{align}
    \hat{y}_{i,j,d} = \text{MLP}^{\text{Output}}_i(\hat{\mathbf{h}}_{i,j,d}).
    \label{Eq:mlp}
\end{align}

To update the local model parameters, gradient descent is employed following the loss value calculated between the predicted values and the ground truth. For $\hat{y}_{i,j,d}$, we adopt cross-entropy, one of the most classical loss functions for data-driven classification problems, to calculate the loss:
\begin{align}
    L^{\text{CE}}_{i,j,d} = \mathbb{L}^{\text{CE}}(\hat{y}_{i,j,d},y_{i,j,d}),
    \label{Eq:ce}
\end{align}
where $\mathbb{L}^{\text{CE}}$ is the cross-entropy loss function.

However, cross-entropy alone may not be sufficiently effective for the local model training in each community. In Section~\ref{sec:fl}, we will introduce another loss function and integrate it with the cross-entropy loss function to collaboratively enhance local model updates.

\subsection{Federated Learning Framework for PV Generation Fraud Detection}\label{sec:fl}

\subsubsection{Model Splitting for Knowledge Sharing in FL}

Since limited data in individual communities hinders local model performance, knowledge sharing for local models through FL communication is essential.
To facilitate knowledge sharing, we adopt a model splitting mechanism that divides each community's local model $f_i(\theta_i)$ into a base model (shallow layers of a DL model) and a head model (deep layers of a DL model). 
The base model, which learns more general features than the head model~\cite{NIPS2014_375c7134}, is uploaded to the server.
Meanwhile, the head model, which specializes in PVG-FD classification, remains locally.

Assume there are $R$ rounds for the FL communication. At round $r$, the local model of community $i$ is denoted as $\theta_{i,r}$. For model splitting, the local model is divided into two components:
\begin{enumerate}
    \item \textbf{Base Model} $\theta^{\text{Base}}_{i,r}$: This component processes data through LSTM, CNN-LSTM, and co-attention layers, capturing generalized features shared across all classes. The base model parameters are represented as a vector in $\mathbb{R}^{P^{\text{Base}}}$, where $P^{\text{Base}}$ denotes the number of learnable base parameters.
    \item \textbf{Head Model} $\theta^{\text{Head}}_{i,r}$: This component consists of the MLP layer, which specializes in refining class-specific decision boundaries. The head model parameters are represented as a vector in $\mathbb{R}^{P^{\text{Head}}}$, where $P^{\text{Head}}$ denotes the number of head parameters.
\end{enumerate}

The model splitting can be denoted as:
\begin{align}
    \theta_{i,r} = [\theta^{\text{Base}}_{i,r}; \theta^{\text{Head}}_{i,r}].
\end{align}

After each community uploads the local base model, the cloud server aggregates these base models using a weighted averaging approach based on data volumes owned by the corresponding local datasets:
\begin{align}
    \theta^{\text{Base},\text{Global}}_r = \sum_{i=1}^{N} \frac{|\mathcal{D}_i|}{|\mathcal{D}|} \theta^{\text{Base}}_{i,r}.
\end{align}
Finally, the cloud server sends the global base model to each community for concatenation with its local head model:
\begin{align}
    \theta_{i,r+1} = [\theta^{\text{Base},\text{Global}}_{r}; \theta^{\text{Head}}_{i,r}].
    \label{Eq:concatenation}
\end{align}

\subsubsection{Prototype-Based Regularization for Imbalanced Data}

To mitigate the impact of data imbalance caused by scarce generation fraud samples on model learning, we adopt prototypes in representation learning, where samples of the same class are clustered around a representative prototype in the learned embedding space~\cite{NIPS2017_cb8da676}. The prototypes are constructed by aggregating the representations of the samples of the corresponding class, which are extracted from the local base model.

At round $r$, for the $i$-th community, assume $\mathcal{K}$ denotes the set of all classes across all the local datasets, i.e., the normal type and all fraud types occurring in $\{\mathcal{Y}_i\}_{i=1}^N$.
Accordingly, the local prototype for class $k \in \mathcal{K}$ is formulated as:
\begin{align}
    \mathbf{p}_{i,k,r} = \frac{1}{|\mathcal{D}_{i,k}|} \sum_{(j,d)\in\mathcal{D}_{i,k}} \hat{\mathbf{h}}_{i,j,d,r} \in \mathbb{R}^{D^{p}},
    \label{Eq:localp}
\end{align}
where $\mathcal{D}_{i,k} \subseteq \mathcal{D}_{i}$ represents the set of all samples in the $i$-th community that belong to class $k$, and $|\mathcal{D}_{i,k}|$ is the total number of such samples.

After receiving the local prototypes from all communities, the global prototypes are computed through weighted aggregation to reflect the contribution of each community proportionally to its data size. 
For class $k$, the global prototype is formulated as:
\begin{align}
    \mathbf{p}_{k,r}^{\text{Global}} = \sum_{i=1}^N\frac{|\mathcal{D}_{i,k}|\cdot\mathbf{p}_{i,k,r}}{\sum_{i=1}^N|\mathcal{D}_{i,k}|} \in \mathbb{R}^{D^{p}}.
    \label{Eq:globalp}
\end{align}

Next, a regularization term is introduced into the local loss function that aligns local prototypes with global prototypes. This alignment enhances the representation quality of minority classes by better embedding their features into the global knowledge.

At round $r$, for the $i$-th community, the regularization term is defined as:
\begin{align}
L^\text{Reg}_{i,r} &= \mathbb{L}^\text{Reg} (\mathbf{p}_{i,r},\mathbf{p}_{r}^{\text{Global}}) \notag\\
&=\sum_{k\in\mathcal{K}} \big[1-\text{cos}(\mathbf{p}_{i,k,r},\mathbf{p}_{k,r}^{\text{Global}})\big],
\label{Eq:regloss}
\end{align}
where $\text{cos}(\cdot)$ denotes cosine similarity between two vectors.
Cosine similarity measures the angular distance between two vectors, focusing on their direction rather than magnitude. This property is particularly beneficial for minority class prototypes, as the small number of samples can lead to unstable magnitudes. By using cosine similarity, the alignment process prioritizes the consistency of feature directions, which is critical for improving minority class representation.

This regularization term is added to the total loss function of the local model as follows:
\begin{align}
L_{i,r} = \sum_{j=1}^{M_i}\sum_{d=1}^{D}L_{i,j,d,r}^{\text{CE}} + \lambda L^\text{Reg}_{i,r},
\label{Eq:totalloss}
\end{align}
where $\lambda$ is a hyperparameter controlling the balance between the cross-entropy loss and the regularization term.

\subsubsection{Training Process for Distributed PVG-FD}

\begin{figure}[t]
\centering
\includegraphics[width=0.49\textwidth]{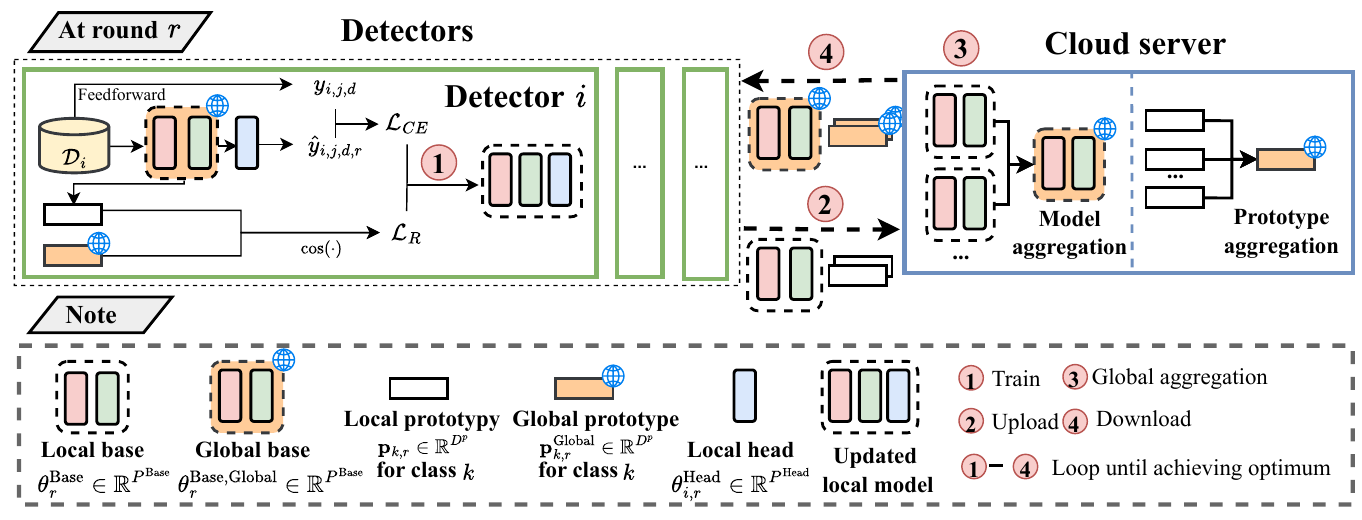}
\caption{Overview of the FL protocol in the proposed PVG-FD framework.}
\label{fig:crop_FL_diagram}
\end{figure}

As illustrated in Fig.~\ref{fig:crop_FL_diagram}, the proposed federated training proceeds in iterative communication rounds between the communities and the cloud server. 
At the beginning of the FL communication, each community trains its own local model, splits the local model into a base and a head, and generates the local class-specific prototypes. 
Then, each community uploads the base model parameters $\theta^{\text{Base}}_{i,r} \in \mathbb{R}^{P^{\text{Base}}}$ and its local class-specific prototypes $\{\mathbf{p}_{i,k,r} \in \mathbb{R}^{D^p}\}_{k=1}^{K}$ to the cloud server for global aggregation. 
After global aggregation, the cloud server sends the global base model parameters  $\theta^{\text{Base,Global}}_{r}\in \mathbb{R}^{P^{\text{Base}}}$ and the global class-specific prototypes $\{\mathbf{p}^{Global}_{k,r}\in \mathbb{R}^{D^p}\}_{k=1}^{K}$ back to each community for local training. 
Finally, the global base and the original local head are concatenated to form a new local model and updated by local training. 
This procedure is repeated until convergence.

Additionally, Algorithm~\ref{alg1} outlines the training process of the privacy-preserving distributed PVG-FD framework. The process consists of three main stages: (1) Initialization, where local models are trained at each community and the global base model is initialized at the server (Lines 4-5); (2) FL communication rounds, in which the communities and the server exchange model parameters and prototypes (Lines 7-33); and (3) Finalization, in which after all communication rounds are completed, the optimized PVG-FD models $\{\theta^*_i\}_{i=1}^N$ are obtained for all data centers (Line 35).

\renewcommand{\baselinestretch}{1.1} 
\begin{algorithm}[t] \footnotesize
\caption{Privacy-Preserving Distributed PVG-FD Method}\label{alg1}
\begin{algorithmic}[1]
\STATE {\bfseries Input:} Local data $\{\mathcal{D}_i\}_{i=1}^N$; Local models $\{\theta_i\}_{i=1}^N$; Learning rates $\{\eta_i\}_{i=1}^N$; Number of communication rounds $R$. 
\STATE {\bfseries Output:} Optimized PVG-FD models $\{\theta^*_i\}_{i=1}^N$. 

\STATE {\bfseries Initialization:}
\STATE Each community $i$ initializes its model $\theta_{i,0}$. 
\STATE Server initializes the global base model $\theta^{\text{Base,Global}}_{0}$. 

\STATE {\bfseries FL communication:}
\STATE \textbf{for} communication round $r = 1$ to $R$ \textbf{do}
\STATE \hspace{0.3cm} {\bfseries Clients:}
\STATE \hspace{0.3cm} \textbf{for} each community $i$ in parallel \textbf{do} \hfill $\blacktriangleright$ \textbf{Local Model Update}

\STATE \hspace{0.6cm} Obtain concatenated local model $\theta_{i,r}$ according to Eq.~(\ref{Eq:concatenation}). 

\STATE \hspace{0.6cm} \textbf{for} each prosumer $j$ \textbf{do} \hfill $\blacktriangleright$ \textbf{Local Training}
\STATE \hspace{0.9cm} \textbf{for} each day $d$ \textbf{do}
\STATE \hspace{1.2cm} Obtain input $\mathbf{X}_{i,j,d}^{\text{PVG}}$ and $\mathbf{X}_{i,j,d}^{\text{Irr}}$.
\STATE \hspace{1.2cm} Compute embeddings $\mathbf{H}^{\text{PVG}}_{i,j,d}$ and $\mathbf{H}^{\text{Irr}}_{i,j,d}$ according to Eq.~(\ref{Eq:lstm}) and Eq.~(\ref{Eq:cnn-lstm}).
\STATE \hspace{1.2cm} Obtain fused embedding $\hat{\mathbf{h}}_{i,j,d}$ according to Eq.~(\ref{Eq:flatten}).
\STATE \hspace{1.2cm} Predict PV generation fraud $\hat{y}_{i,j,d}$ according to Eq.~(\ref{Eq:mlp}).
\STATE \hspace{0.9cm} \textbf{end for}
\STATE \hspace{0.6cm} \textbf{end for}

\STATE \hspace{0.6cm} \textbf{if} $r > 1$ \textbf{then}
\STATE \hspace{0.9cm} Compute regularization term $L^{\text{Reg}}_{i,r}(\theta_{i,r})$ according to Eq.~(\ref{Eq:regloss}).
\STATE \hspace{0.6cm} \textbf{else}
\STATE \hspace{0.9cm} Set $L^{\text{Reg}}_{i,r}(\theta_{i,r}) = 0$.
\STATE \hspace{0.6cm} \textbf{end if}

\STATE \hspace{0.6cm} Compute total loss $L_{i,r}(\theta_{i,r})$ using Eq.~(\ref{Eq:totalloss}). 
\STATE \hspace{0.6cm} Update model parameters $\theta_{i,r} \leftarrow \theta_{i,r} - \eta_i \nabla_{\theta_{i,r}} L_{i,r}(\theta_{i,r})$.

\STATE \hspace{0.6cm} Compute local class-specific prototypes $\{\mathbf{p}_{i,k,r}\}_{k\in\mathcal{K}}$ using Eq.~(\ref{Eq:localp}). 
\STATE \hspace{0.6cm} Split model into base $\theta^{\text{Base}}_{i,r}$ and head $\theta^{\text{Head}}_{i,r}$. 
\STATE \hspace{0.6cm} Send $\theta^{\text{Base}}_{i,r}$ and $\{\mathbf{p}_{i,k,r}\}_{k\in\mathcal{K}}$ to the server. 

\STATE \hspace{0.3cm} \textbf{end for}

\STATE \hspace{0.3cm} {\bfseries Server:} \hfill $\blacktriangleright$ \textbf{Global Aggregation}
\STATE \hspace{0.3cm} Aggregate base models to obtain global base model $\theta^{\text{Base,Global}}_{r}$.
\STATE \hspace{0.3cm} Aggregate local class-specific prototypes to obtain $\{\mathbf{p}^{\text{Global}}_{k,r}\}_{k\in\mathcal{K}}$.
\STATE \hspace{0.3cm} Send $\theta^{\text{Base,Global}}_{r}$ and $\{\mathbf{p}^{\text{Global}}_{k,r}\}_{k\in\mathcal{K}}$ to the clients.

\STATE \textbf{end for}

\STATE \textbf{return} $\theta^*_1, \theta^*_2, \dots, \theta^*_N$. 
\end{algorithmic}
\end{algorithm}
\renewcommand{\baselinestretch}{1.0}

\section{Experimental Study and Results}\label{Sec:E}

In this section, we conduct experiments to evaluate the effectiveness of our developed method. We begin with a description of the dataset used in our experiments, followed by a description of the performance metrics and baseline methods for comparison. Finally, we analyze the results, focusing on the performance of local models, the effectiveness of the FL framework, and the robustness of our approach against class imbalance.

\subsection{Dataset Description}

\begin{figure}[t]
\centering
\includegraphics[width=0.48\textwidth]{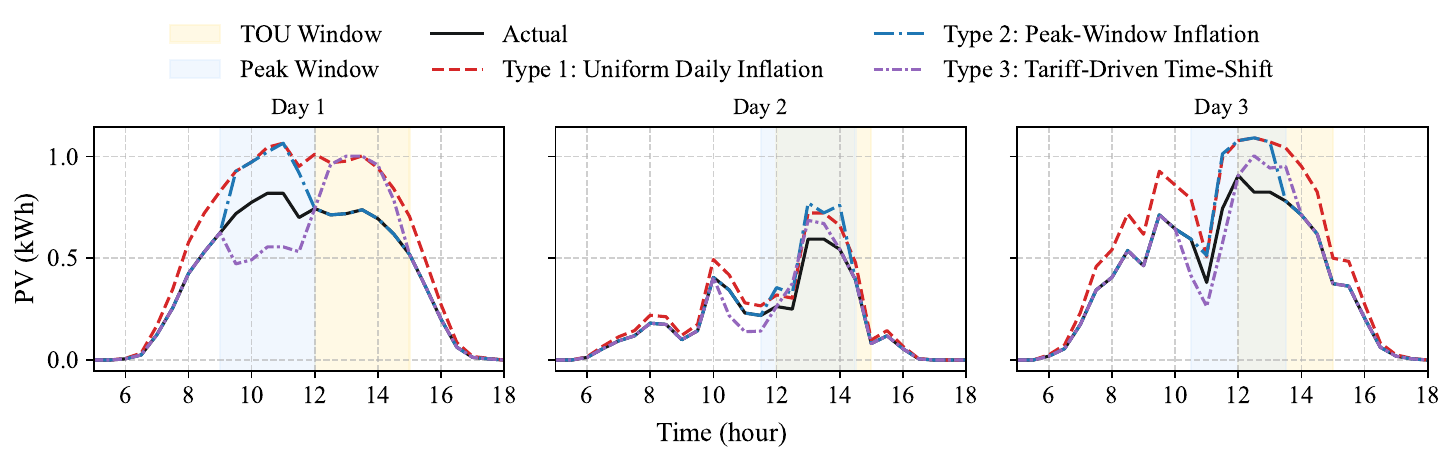}
\caption{Comparison of normal and malicious solar generation data for customer ID 239 over a 72-hour period (September 23–25, 2011).}
\label{fig:theft_ana}
\end{figure}

The dataset employed in these experiments is the Solar Home Electricity Data~\cite{ratnam2017residential} provided by Ausgrid's electricity network, encompassing three years of half-hourly electricity data for 300 randomly selected homes with rooftop PV systems. The data collection period spans from July 1, 2010, to June 30, 2013, gathered through meter reading processes, thus covering multiple seasonal cycles and capturing a wide range of operational conditions. Notably, the dataset records gross generation, measuring the total electricity output from the PV systems independent of actual residential consumption.
In addition to electricity data, comprehensive meteorological records for the same period are obtained from the National Solar Radiation Database~\cite{NREL}. This database provides critical weather parameters needed to assess solar irradiance conditions accurately. In particular, this study focuses on three principal solar radiation metrics: GHI, DNI, and DHI, which collectively characterize the solar resource.

In the FL setting, the complete dataset is first partitioned into multiple local datasets, each corresponding to an independent community. The partitioning is performed at the prosumer level, such that each prosumer belongs to exactly one local dataset. In the experiments, the default number of communities is set to $N = 5$, with each community consisting of approximately 60 different prosumers.
In each local dataset, the data are further divided into training and testing sets in temporal order. Due to the unavailability of meteorological variables in the first six months, the effective data span covers approximately 2.5 years. The first full year is used as the training period, while the test set is constructed only from the subsequent year. To maintain a commonly used training proportion of approximately 75\%, one-third of the days in the second year are randomly sampled at the day level to form the test set.

Since the original dataset is derived from prosumers who have installed smart meters and voluntarily agreed to share their data, it is reasonable to assume that all samples in this dataset correspond to honest prosumers. 
Following prior works~\cite{7108042,8998142,TANG2023109212}, malicious samples are synthetically generated from benign ones by applying PV generation fraud functions. 
Specifically, a randomly selected subset of benign prosumer-day samples is converted into malicious samples by applying one of the fraud functions to each selected PV generation profile, while the remaining samples are kept unchanged.
In this study, we consider three representative types of PV generation fraud, each capturing a distinct manipulation behavior. Type~1 models a uniform, shape-preserving inflation, where the reported PV generation is proportionally increased across the entire day. Type~2 concentrates the inflation within a peak-generation window, resulting in localized amplification around high-irradiance periods. Type~3 represents a tariff-driven time-shift attack, in which part of the daytime PV generation is redistributed toward peak-tariff hours under a daily energy-conservation constraint.
Fig.~\ref{fig:theft_ana} illustrates some examples of the three PV generation fraud types over three consecutive days, showing their distinct manipulation patterns in comparison with the actual PV generation profiles.
The three fraud strategies are formulated as follows.

Type~1 applies a uniform inflation to the reported PV generation across the entire day. The manipulated PV generation is defined as
\begin{equation}
    x^{\text{PVG}}_{i,d,t} = \min\!\left(
        x^{\text{APVG}}_{i,d,t} (1+\alpha_{d}),
        \; x^{\max}_{i,t}
    \right),
\end{equation}
where $\alpha_d$ is a random inflation factor controlling the daily inflation magnitude, and $x^{\max}_{i,t}$ denotes a data-driven instantaneous upper bound at time slot $t$ (enlarged by a small safety margin).

Type~2 localizes the inflation to a peak-generation window rather than applying it uniformly over the day.
Let
\begin{equation}
    t^{\text{peak}}_{i,d}
    =
    \arg\max_{t} x^{\text{APVG}}_{i,d,t}
\end{equation}
denote the peak PV generation time slot of day $d$ (taking the earliest one if multiple peaks exist), and define the peak-centered window: 
\begin{equation}
    \mathcal{T}^{\text{peak}}_{i,d} = \{t : |t - t^{\text{peak}}_{i,d}| < K^{\text{Win}}\},
\end{equation}
where $K^{\text{Win}}$ is a predefined half-window length (in time slots). 
The reported PV generation is inflated only in this window:
\begin{equation}
    x^{\text{PVG}}_{i,d,t}
    =
    \begin{cases}
        \min\!\left(
            x^{\text{APVG}}_{i,d,t}(1+\beta_{i,d,t}),\;
            x^{\max}_{i,t}
        \right),
        & t \in \mathcal{T}^{\text{peak}}_{i,d},
        \\[6pt]
        x^{\text{APVG}}_{i,d,t},
        & t \notin \mathcal{T}^{\text{peak}}_{i,d},
    \end{cases}
\end{equation}
where $\beta_{i,d,t}$ controls the slot-wise inflation in peak windows.

Unlike Types~1 and~2, which manipulate generation magnitudes, Type~3 redistributes PV generation across time slots under a daily energy-conservation constraint.
Let the daylight window be
\begin{equation}
    \mathcal{T}^{\text{sun}}_{i,d} = \{t : x^{\text{APVG}}_{i,d,t} > \tau\},
\end{equation}
and let $\mathcal{T}^{\text{TOU}}$ denote the set of peak-tariff time-slot indices.
The attacker selects a source set $\mathcal{S}_{i,d} \subset \mathcal{T}^{\text{sun}}_{i,d} \setminus \mathcal{T}^{\text{TOU}}$ such that $t+\Delta_{i,d}\in\mathcal{T}^{\text{TOU}}$ for all $t\in\mathcal{S}_{i,d}$, and shifts a fraction of its energy forward by $\Delta_{i,d}$.
For each $t \in \mathcal{S}_{i,d}$, the shifted target slot is $u=t+\Delta_{i,d}$.
Let $\lambda_{i,d,t}$ denote the fraction of energy 
moved from slot $t$.
The shift length $\Delta_{i,d}$ and the energy transfer ratio $\lambda_{i,d,t}$ are sampled from predefined ranges.
The pre-clipping manipulated profile $\tilde{x}^{\text{PVG}}_{i,d,t}$ is given by
\begin{equation}
\tilde{x}^{\text{PVG}}_{i,d,t}
=
\begin{cases}
(1-\lambda_{i,d,t})\,x^{\text{APVG}}_{i,d,t}, 
& t \in \mathcal{S}_{i,d}, \\[6pt]
x^{\text{APVG}}_{i,d,t} + 
\displaystyle\sum_{\substack{s \in \mathcal{S}_{i,d}:\\ s+\Delta_{i,d}=t}}
\lambda_{i,d,s}\,x^{\text{APVG}}_{i,d,s},
& t \in \mathcal{T}^{\text{TOU}}, \\
x^{\text{APVG}}_{i,d,t}, 
& \text{otherwise}.
\end{cases}
\end{equation}
The redistribution preserves the total daily PV generation before clipping.
Finally, the reported PV generation is clipped by the instantaneous upper bound $x^{\max}_{i,t}$, i.e., $x^{\text{PVG}}_{i,d,t} = \min\!\left(\tilde{x}^{\text{PVG}}_{i,d,t},\, x^{\max}_{i,t}\right)$.

In the default experimental setting, fraud samples constitute 15\% of the dataset, evenly distributed across the three fraud types.

\subsection{Performance Metrics and Benchmark Settings}
To evaluate the performance of the proposed framework, we employ four evaluation metrics: accuracy, Area Under the Curve (AUC), F1-score, and Matthews Correlation Coefficient (MCC). 

In the proposed method, we integrate a novel local model and an FL framework.
The local detector deployed at each community is a residential-level PVG-FD model that determines whether the reported PV generation for a given day is fraudulent. All communities employ the same model architecture and hyperparameter settings, while the learned parameters differ due to community-specific data.
To comprehensively evaluate the performance of each component, we separately compare the local model with baseline deep learning models and the FL framework with established FL methods. 
For the local model, we conduct distributed evaluations under the FedAvg~\cite{pmlr-v54-mcmahan17a} framework and compare it against several deep learning baselines, including LSTM (a two-layer LSTM with the final hidden state used as the sequence representation)~\cite{hochreiter1997long}, CNN-LSTM (two one-dimensional convolutional layers followed by a two-layer LSTM)~\cite{Donahue_2015_CVPR}, 
Transformer~\cite{10.5555/3295222.3295349}, 
Reformer~\cite{Kitaev2020Reformer}, 
and DLinear~\cite{Zeng_Chen_Zhang_Xu_2023}.
For Transformer, Reformer, and DLinear, we adopt the standard architectures and recommended configurations from the original papers.
For the FL framework, we benchmark its performance against four baselines: Local-only, FedAvg~\cite{pmlr-v54-mcmahan17a} (a traditional FL framework), BalanceFL~\cite{9825928}, and FedNH~\cite{Dai_Chen_Li_Heinecke_Sun_Xu_2023} (FL frameworks designed to address class imbalance). In the Local-only approach, each data center independently trains a model using only its local data, serving as a baseline for fully decentralized learning. When evaluating FL frameworks, we implement all frameworks using the proposed local model for a consistent and fair comparison.

The hyperparameters used in this study, including those associated with the methodology, experimental setup, and training configuration, are summarized in Table~\ref{tab:hyperparameters}.

\begin{table}[h!]
\centering

\setlength{\tabcolsep}{4pt}
\caption{Summary of hyperparameters used in the proposed PVG-FD method and experimental settings.}
\label{tab:hyperparameters}
\begin{tabular}{ll}
\hline\hline
\textbf{Component} & \textbf{Hyperparameter} \\ 
\hline
\multicolumn{2}{l}{\textbf{Methodology}} \\
\quad Sliding window length $T^{\text{Sub}}$ & 24\\
\quad Sliding window stride $S$ & 6 \\[2pt]
\quad Hidden size $d_{\text{LSTM}}$ & 128 \\[2pt]
\quad Hidden size $d_{\text{CNN-LSTM}}$ & 128 \\[2pt]
\quad Self-attention dimension $d_{\text{SA}}$ & 128 \\
\quad Cross-attention dimension $d_{\text{CA}}$ & 128 \\
\quad Head number $h$ & 4 \\
\quad Regularization weight $\lambda$ & 1\\[2pt]

\multicolumn{2}{l}{\textbf{Experiment}} \\
\quad Number of daily time slots $T$ & 48 \\
\quad Default number of communities $N$ & 5 \\
\quad Daily inflation factor $\alpha_d$ & Uniform(0.1,0.5) \\
\quad Peak-window inflation factor $\beta_{i,d,t}$ & Uniform(0.1,0.5) \\
\quad Peak-window half width $K^{\text{Win}}$ & 3 \\
\quad Energy transfer ratio $\lambda_{i,d,t}$ & Uniform(0.1,0.5) \\
\quad Time-shift length $\Delta_{i,d}$ & $\{4,5,6\}$ \\
\quad Peak-tariff window $\mathcal{T}^{\text{TOU}}$ & 12:00-15:00 \\
\quad FL rounds $R$ & 100 \\
\quad Learning rate $\eta$ & $1\times10^{-4}$ \\
\quad Local batch size & 64 \\
\quad Local update steps per round & 1 \\[2pt]
\hline\hline
\end{tabular}
\end{table}

\subsection{Result Analysis}

\subsubsection{Comparison of Local Model Performance}

\begin{table}[]
\centering
\renewcommand{\arraystretch}{1.2} 
\setlength{\tabcolsep}{3.2pt} 
\scriptsize
\caption{Comparative Evaluation of Local Model Performance under the FedAvg Framework}
\label{tab:table1}
\begin{tabular}{ccccc}
\hline\hline
Local Method   & Acc.±Std.(\%)       & AUC±Std.(\%)        & F1±Std.(\%)         & MCC±Std.(\%)        \\ \hline
LSTM           & 92.91±0.50          & 92.52±0.45          & 73.92±2.35          & 70.83±2.56          \\
CNN-LSTM       & 93.36±0.34          & 93.82±0.43          & 74.91±1.94          & 72.60±1.91          \\
Transformer    & 94.82±0.40          & 95.52±0.24          & 81.57±1.83          & 79.13±1.90          \\
Reformer       & 94.46±0.24          & 95.77±0.29          & 80.04±2.03          & 77.55±1.31          \\
DLinear        & 93.89±0.30          & 95.00±0.37          & 77.63±1.98          & 75.07±2.01          \\
Proposed       & \textbf{95.57±0.29} & \textbf{96.58±0.23} & \textbf{84.17±1.48} & \textbf{82.26±1.46}  \\
\hline\hline
\end{tabular}
\end{table}

\begin{figure}[ht]
    \centering   
    \includegraphics[width=.3\textwidth]{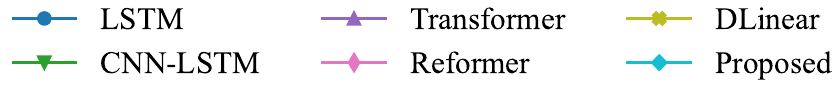}
    \subfigure{
        \includegraphics[width=.22\textwidth]{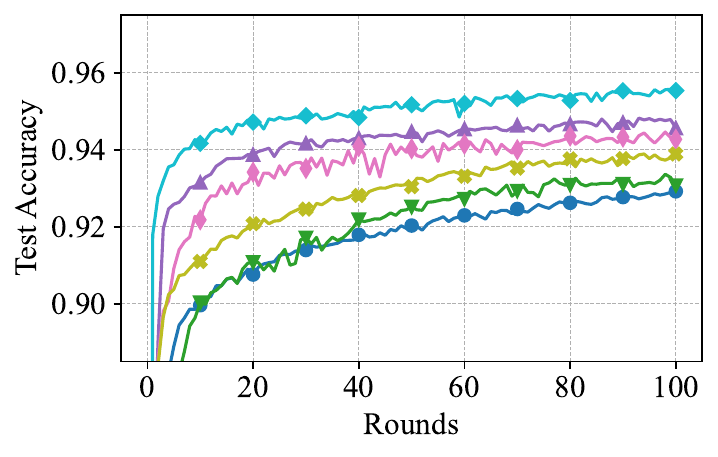}  
        \label{Fig:curve1_acc_updated}
        }
    \subfigure{
        \includegraphics[width=.22\textwidth]{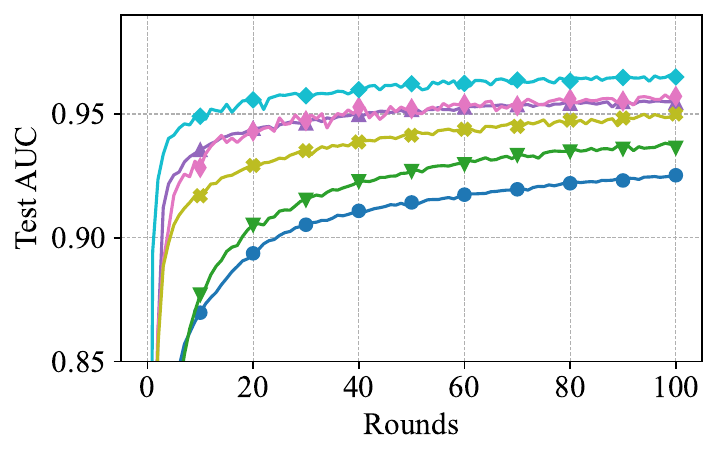} 
        \label{Fig:curve1_auc_updated}
        }
    \subfigure{
        \includegraphics[width=.22\textwidth]{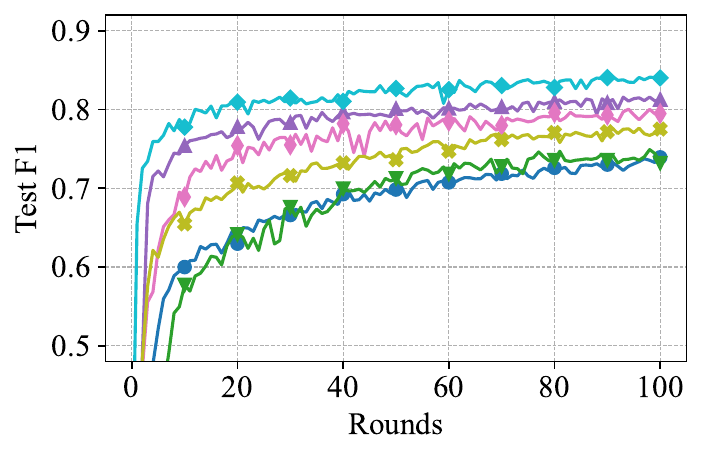}  
        \label{Fig:curve1_f1_updated}
        }
    \subfigure{
        \includegraphics[width=.22\textwidth]{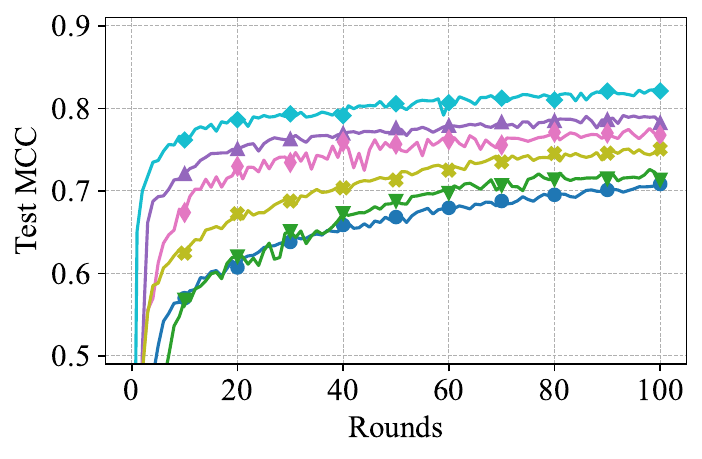}  
        \label{Fig:curve1_mcc_updated}
        }
    \caption{Learning curves of local models under the FedAvg framework across four evaluation metrics.}
    \label{Fig:all_curve1}
\end{figure}

\begin{table*}[h!pt]
\setlength{\tabcolsep}{6pt}
\renewcommand{\arraystretch}{1.2} 
\centering
\caption{Performance of local detection models under FedAvg when trained on three seasons in Year~1 and tested on the held-out target season in Year~2.}
\label{tab:season}
\resizebox{\textwidth}{!}{
\begin{tabular}{ccccccccccccccccc}
\hline\hline
\multirow{2}{*}{Local Method}             & \multicolumn{4}{c}{Spring (\%)}                                        & \multicolumn{4}{c}{Summer (\%)}                                        & \multicolumn{4}{c}{Autumn (\%)}                                        & \multicolumn{4}{c}{Winter (\%)}                                        \\\cline{2-17}
             & Acc.           & AUC            & F1             & MCC            & Acc.           & AUC            & F1             & MCC            & Acc.           & AUC            & F1             & MCC            & Acc.           & AUC            & F1             & MCC            \\ \hline
LSTM         & 92.22          & 90.21          & 71.02          & 67.72          & 74.40          & 79.96          & 43.57          & 31.76          & 92.29          & 90.72          & 71.23          & 68.00          & 88.08          & 72.27          & 45.28          & 45.01          \\
CNN-LSTM     & 92.32          & 92.33          & 71.74          & 68.30          & 75.82          & 81.86          & 45.48          & 34.27          & 92.32          & 92.11          & 71.50          & 68.20          & 88.75          & 77.83          & 48.57          & 48.86          \\
Transformer  & 93.78          & 94.25          & 78.42          & 75.10          & 79.33          & 87.37          & 52.89          & 44.19          & 93.81          & 94.11          & 78.41          & 75.16          & 91.28          & 78.90          & 63.05          & 62.56          \\
Reformer     & 93.42          & 94.42          & 76.80          & 73.43          & 78.24          & 85.97          & 50.53          & 41.04          & 93.48          & 94.28          & 76.98          & 73.67          & 90.28          & 77.94          & 58.18          & 57.41          \\
DLinear      & 92.74          & 93.23          & 74.01          & 70.39          & 76.72          & 82.60          & 47.93          & 37.63          & 92.79          & 93.42          & 74.19          & 70.62          & 89.81          & 78.32          & 54.38          & 54.82          \\
Proposed     & \textbf{94.72} & \textbf{96.04} & \textbf{81.47} & \textbf{78.83} & \textbf{80.02} & \textbf{89.27} & \textbf{54.16} & \textbf{45.84} & \textbf{94.77} & \textbf{96.09} & \textbf{81.58} & \textbf{79.00} & \textbf{92.12} & \textbf{80.38} & \textbf{67.50} & \textbf{66.73} \\ \hline\hline
\end{tabular}}
\end{table*}

Table~\ref{tab:table1} compares local model performance aggregated across all participating communities under the FedAvg framework. For each method, the mean and standard deviation of accuracy, AUC, F1-score, and MCC are reported to capture both central tendency and variability.
The proposed local model achieves the highest mean values in all metrics, along with relatively small standard deviations, demonstrating superior performance.
Among the baseline methods, the self-attention-based Transformer architectures (Transformer and Reformer) exhibit competitive performance. The Reformer achieves a slightly higher AUC than the Transformer, while the Transformer attains marginally superior accuracy, F1-score, and MCC. In contrast, DLinear achieves comparable accuracy and F1-scores but exhibits weaker performance in AUC and MCC. LSTM and CNN-LSTM show the weakest performance, indicating limited effectiveness in handling multi-source inputs and complex temporal patterns.
Fig.~\ref{Fig:all_curve1} presents the learning curves of different local models under the FedAvg framework across four evaluation metrics, illustrating their performance evolution over 100 communication rounds.
Overall, the models show a clear upward learning trend and gradually converge to stable performance levels, with moderate round-to-round fluctuations typically caused by mini-batch optimization.
Among the compared methods, the proposed model consistently achieves higher performance throughout the training process and converges to a stable level within the early communication rounds across all metrics. In comparison, Transformer and Reformer exhibit similar but slightly slower convergence behaviors, while DLinear shows a more gradual performance improvement. LSTM and CNN-LSTM converge more slowly and reach lower final performance levels, which is consistent with the results reported in Table~\ref{tab:table1}.

Moreover, we further examine the performance of local detection models under a cross-season evaluation setting, where models are trained on data from three seasons in Year~1 and evaluated on the held-out target season in Year~2 under the FedAvg framework. The corresponding results are summarized in Table~\ref{tab:season}.
Compared with the standard data split, all models exhibit varying degrees of performance degradation across seasons. When spring or autumn is used as the target season, the degradation is relatively moderate, whereas more pronounced drops are observed for summer and winter. In particular, performance reductions in summer and winter are more evident in F1-score, MCC, and AUC, indicating increased detection difficulty under stronger seasonal variation. Across all four target seasons, the proposed local detection model consistently achieves the best performance among all compared methods.

\begin{table}[t]
\setlength{\tabcolsep}{2.5pt}
\renewcommand{\arraystretch}{1.2} 
\centering
\caption{Performance of FedAvg-based detection models under noisy irradiance conditions.}
\label{tab:noisy_irr}
\begin{tabular}{ccccccccc}
\hline\hline
\multirow{2}{*}{Local Model} & \multicolumn{2}{c}{Acc. (\%)}             & \multicolumn{2}{c}{AUC (\%)}              & \multicolumn{2}{c}{F1 (\%)}               & \multicolumn{2}{c}{MCC (\%)}              \\ \cline{2-9} 
                             & Metric                         & $\Delta$ & Metric                         & $\Delta$ & Metric                         & $\Delta$ & Metric                         & $\Delta$ \\ \hline
LSTM           & 90.49          & -2.42             & 89.84          & -2.68            & 68.63          & -5.29           & 64.21          & -6.62            \\
CNN-LSTM       & 91.20          & -2.16             & 91.37          & -2.45            & 70.37          & -4.54           & 67.55          & -5.05            \\
Transformer    & 91.93          & -2.89             & 92.45          & -3.07            & 73.03          & -8.54           & 70.89          & -8.24            \\
Reformer       & 91.54          & -2.92             & 92.54          & -3.23            & 70.51          & -9.53           & 68.87          & -8.68            \\
DLinear        & 90.92          & -2.97             & 91.84          & -3.16            & 69.90          & -7.73           & 67.38          & -7.69            \\
Proposed       & \textbf{92.49} & -3.08             & \textbf{92.91} & -3.67            & \textbf{74.65} & -9.52           & \textbf{73.43} & -8.83            \\
\hline\hline
\end{tabular}
\end{table}

\subsubsection{Performance under Noisy Irradiance Conditions}

To evaluate the impact of degraded irradiance information on detection performance, we conduct an additional experiment where each sample's irradiance sequence is replaced by the corresponding community-level seasonal average profile. Both training and testing are performed under this noisy irradiance condition using the FedAvg framework, while keeping all other settings unchanged.
The performance results are reported in Table~\ref{tab:noisy_irr}, where each $\Delta$ value denotes the relative change with respect to the main experiment (Table~\ref{tab:table1}) using the original irradiance data. Overall, all local detection models experience noticeable performance degradation under noisy irradiance conditions, reflecting increased challenges in identifying fraudulent cases when fine-grained irradiance information is unavailable.
Despite the degraded input quality, the proposed local detection model consistently achieves the best performance, maintaining a clear advantage over the baseline methods.

\subsubsection{Evaluation of FL Framework Effectiveness}

\begin{table}[]
\centering
\renewcommand{\arraystretch}{1.2} 
\setlength{\tabcolsep}{3.2pt} 
\scriptsize 
\caption{Performance Comparison of FL Frameworks Using the Proposed Local Model}
\label{tab:table2}
\begin{tabular}{ccccc}
\hline\hline
FL Framework   & Acc.±Std.(\%)       & AUC±Std.(\%)        & F1±Std.(\%)         & MCC±Std.(\%)        \\ \hline
Local-only     & 94.70±0.48          & 95.63±0.49          & 81.31±2.01          & 78.69±2.57          \\
FedAvg         & 95.57±0.33          & 96.58±0.37          & 84.17±1.58          & 82.26±1.83          \\
BalanceFL      & 96.13±0.31          & 96.70±0.20          & 86.32±1.45          & 84.62±1.60          \\
FedNH          & 96.43±0.30          & 97.09±0.24          & 87.52±1.47          & 85.88±1.55          \\
Proposed       & \textbf{96.90±0.29} & \textbf{97.49±0.23} & \textbf{89.27±1.32} & \textbf{87.78±1.46}   \\ \hline\hline
\end{tabular}
\end{table}

\begin{figure}[t]
    \centering   
    \includegraphics[width=.48\textwidth]{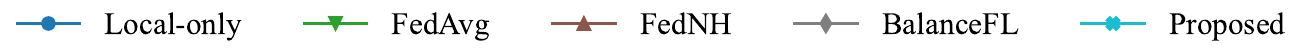}
    \subfigure{
        \includegraphics[width=.22\textwidth]{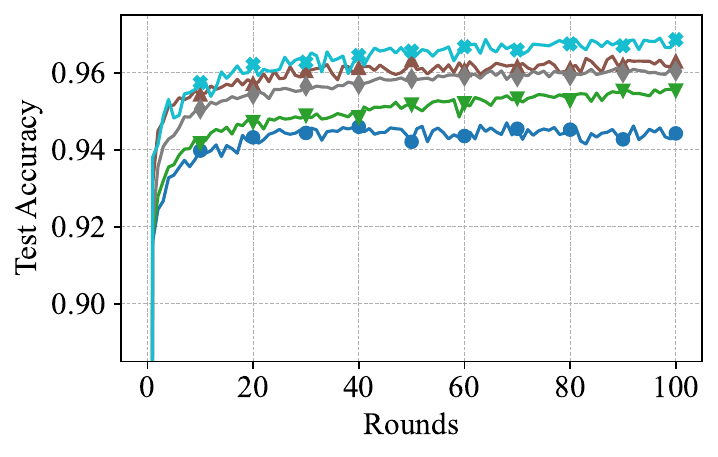}  
        \label{Fig:curve2_acc_updated}
        }
    \subfigure{
        \includegraphics[width=.22\textwidth]{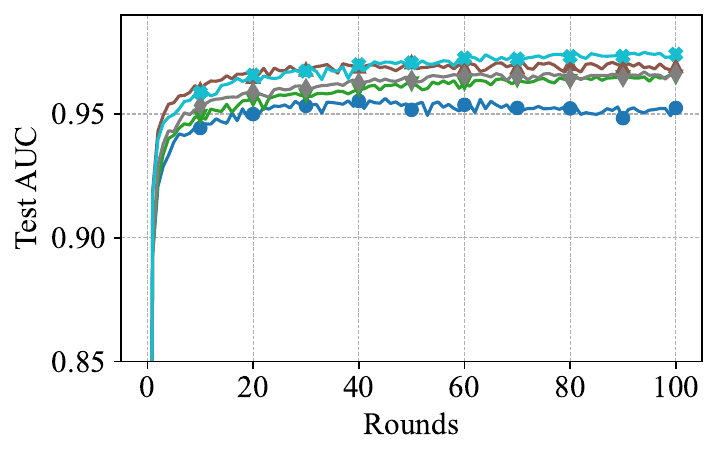} 
        \label{Fig:curve2_auc_updated}
        }
    \subfigure{
        \includegraphics[width=.22\textwidth]{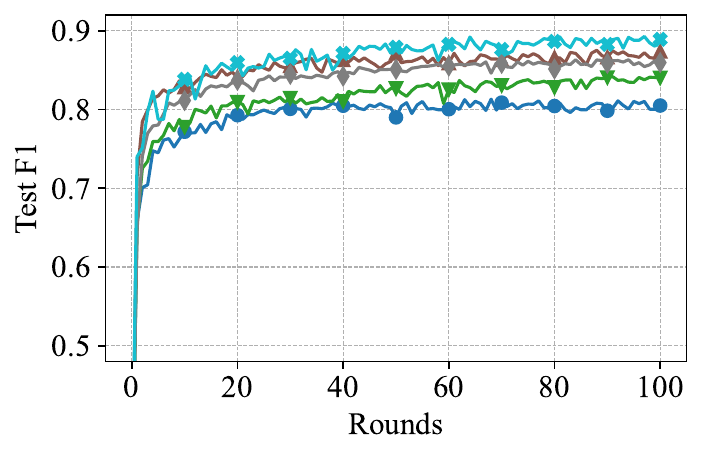}  
        \label{Fig:curve2_f1_updated}
        }
    \subfigure{
        \includegraphics[width=.22\textwidth]{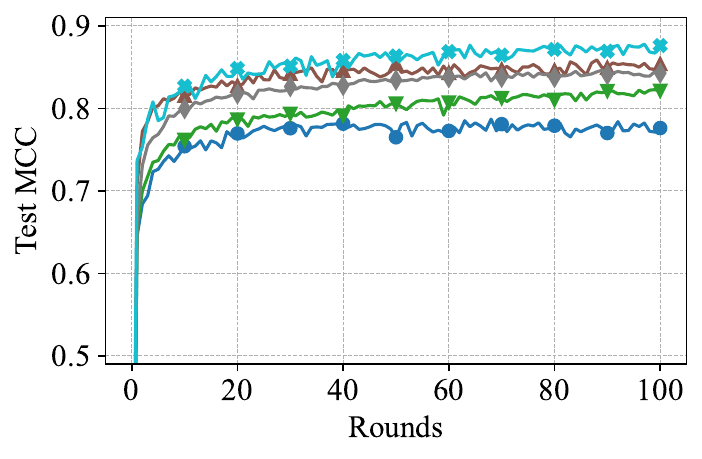}  
        \label{Fig:curve2_mcc_updated}
        }
    \caption{Learning curves of FL frameworks across four evaluation metrics.}
    \label{Fig:all_curve2}
\end{figure}

\begin{figure}[ht]
    \centering   
    \includegraphics[width=.48\textwidth]{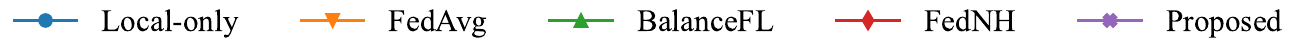}
    \subfigure{
        \includegraphics[width=.22\textwidth]{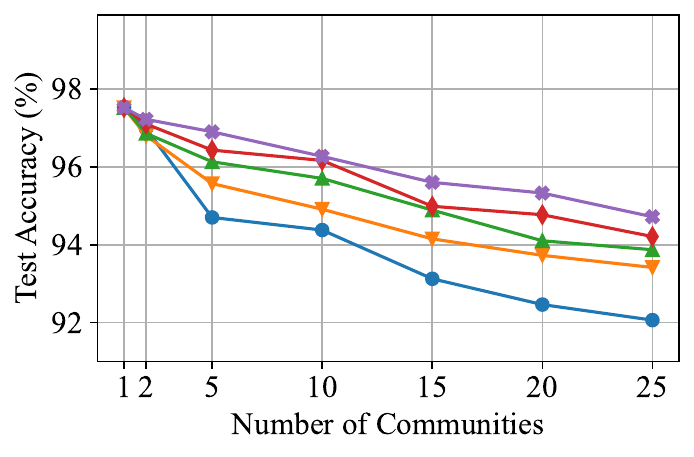}  
        \label{Fig:curve4_acc_R1_y}
        }
    \subfigure{
        \includegraphics[width=.22\textwidth]{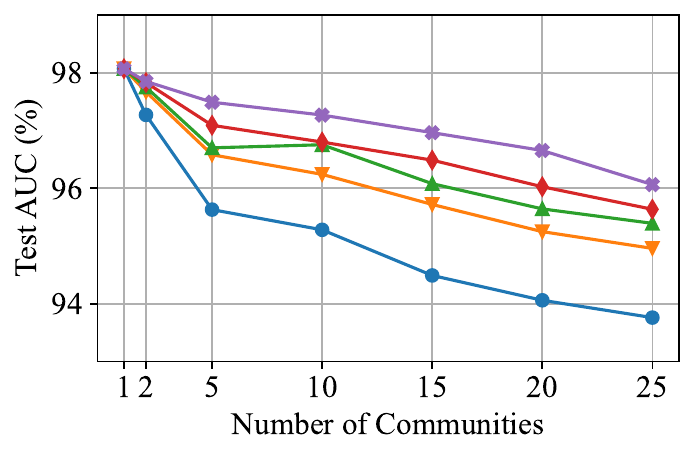} 
        \label{Fig:curve4_auc_R1_y}
        }
    \subfigure{
        \includegraphics[width=.22\textwidth]{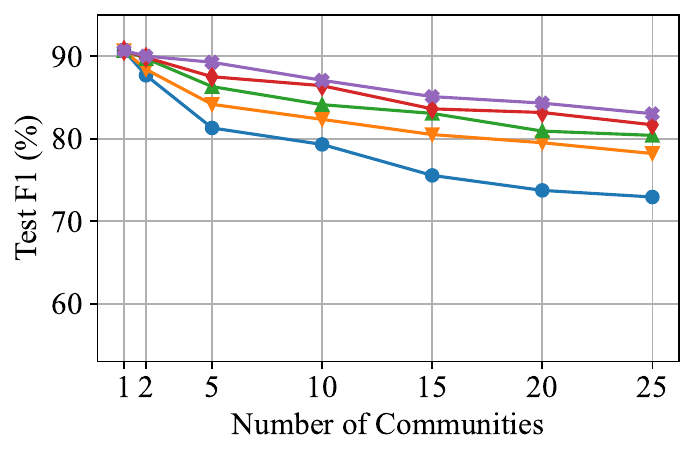}  
        \label{Fig:curve4_f1_R1_y}
        }
    \subfigure{
        \includegraphics[width=.22\textwidth]{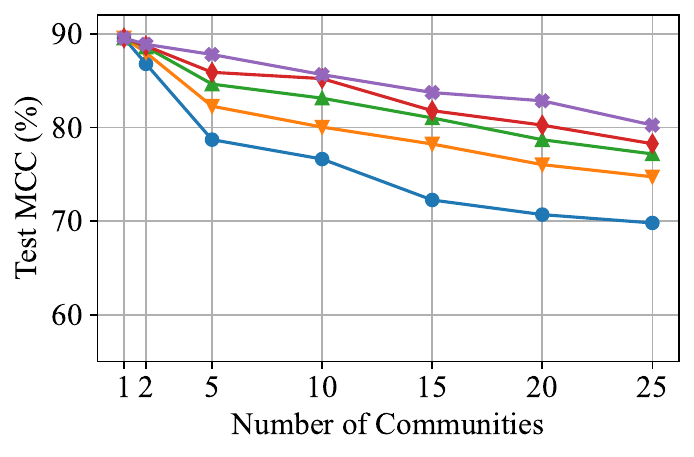}  
        \label{Fig:curve4_mcc_R1_y}
        }
    \caption{Performance comparison under varying community counts with a fixed total of 300 households.}
    \label{Fig:all_curve4}
\end{figure}

Table~\ref{tab:table2} compares the performance of different FL frameworks using the proposed local detection model. Overall, all federated approaches outperform the local-only baseline across all evaluation metrics, showing the benefits of collaborative training in distributed settings.
Among the compared FL frameworks, the proposed framework consistently achieves the best performance in terms of accuracy, AUC, F1-score, and MCC. FedNH and BalanceFL also show strong performance improvements over FedAvg, while FedAvg provides a clear performance gain compared with local-only training. These results indicate that incorporating more advanced FL strategies leads to more effective aggregation and improved detection performance.
Additionally, Fig.~\ref{Fig:all_curve2} illustrates the learning curves of different FL frameworks using the proposed local detection model across four evaluation metrics.
Overall, all federated approaches show rapid performance improvements in the early communication rounds and gradually converge to stable performance levels, while the local-only baseline consistently yields lower performance throughout the training process. Among the compared frameworks, the proposed FL framework maintains the highest performance across all metrics during the training process. FedNH and BalanceFL exhibit similar convergence behaviors and consistently outperform FedAvg, whereas FedAvg shows clear improvements over local-only training. These training dynamics are consistent with the quantitative results reported in Table~\ref{tab:table2}.

We further investigate how the performance of different detection frameworks varies with the number of communities while keeping the total number of households fixed at 300. In this setting, increasing the number of communities only changes the granularity of data partitioning, such that each community contains fewer households, without introducing any additional data.
As shown in Fig.~\ref{Fig:all_curve4}, all methods exhibit a general performance decline as the number of communities increases, which is expected due to reduced local data availability in each community. 
When the number of communities is set to one, the training reduces to a centralized setting. With two communities, the system already operates in a distributed manner, but still benefits from relatively large local data partitions. As the partition becomes finer, the performance of all methods gradually decreases.
Notably, the Local-only baseline shows the most pronounced performance degradation as the number of communities increases, whereas all FL frameworks demonstrate a more gradual decline. This indicates that FL can better mitigate the performance loss caused by reduced local data size by enabling model-level knowledge sharing across communities. Among the federated approaches, the proposed framework consistently maintains the highest performance across different community counts.

\begin{table}[t]
\centering
\caption{Per-round communication overhead comparison of different FL frameworks under the study setting.}
\label{tab:comm_overhead}
\renewcommand{\arraystretch}{1.15}
\begin{tabular}{c c c}
\hline
\textbf{Method} & \textbf{General communication form}                       & \textbf{Number of parameters} \\ \hline
FedAvg          & $2N\left(P^{\text{Base}} + P^{\text{Head}}\right)$        & $9.45 \times 10^6$            \\
BalanceFL       & $2N\left(P^{\text{Base}} + P^{\text{Head}}\right)$        & $9.45 \times 10^6$            \\
FedNH           & $N\left(2P^{\text{Base}} + P^{\text{Head}} + 2D^p\right)$ & $8.52 \times 10^6$            \\
Proposed        & $2N\left(P^{\text{Base}} + 2D^p\right)$                   & $\mathbf{7.60 \times 10^6}$   \\ \hline
\end{tabular}
\end{table}

In addition to detection performance, we further compare different FL frameworks from the perspective of per-round communication overhead, since communication cost is also an important aspect in evaluating the practicality of distributed training. The corresponding results are summarized in Table~\ref{tab:comm_overhead}. Here, $P^{\text{Base}}$ and $P^{\text{Head}}$ denote the numbers of parameters in the base model and the head model of the PV fraud detection model used in this study, respectively, and $D^p$ denotes the dimension of the class-specific prototype. Under the study setting, all $N=5$ communities participate in each communication round. Accordingly, the proposed framework exchanges the base model and two class-specific prototypes in each communication round, while keeping the local head model at the client side. In contrast, FedAvg and BalanceFL exchange the full detection model, whereas FedNH further involves class-wise representation transmission. Based on the parameter sizes of the PV fraud detection model used in this study, the corresponding numerical results are reported in Table~\ref{tab:comm_overhead}. As shown in the table, under the current setting, the proposed framework achieves the lowest per-round communication overhead among the compared FL methods.

\subsubsection{Evaluation under Cross-Community Settings}

\begin{table}[t]
\setlength{\tabcolsep}{2.5pt}
\renewcommand{\arraystretch}{1.2}
\centering
\caption{Performance of federated methods on the held-out community under the few-shot setting.}
\label{tab:cross_com}
\begin{tabular}{ccccccccc}
\hline\hline
\multirow{2}{*}{FL Framework} & \multicolumn{2}{c}{Acc. (\%)}             & \multicolumn{2}{c}{AUC (\%)}              & \multicolumn{2}{c}{F1 (\%)}               & \multicolumn{2}{c}{MCC (\%)}              \\ \cline{2-9} 
                             & Metric                         & $\Delta$ & Metric                         & $\Delta$ & Metric                         & $\Delta$ & Metric                         & $\Delta$ \\ \hline
FedAvg         & 93.93          & -1.64             & 94.29          & -2.29            & 80.28          & -3.89           & 77.62          & -4.64            \\
BalanceFL      & 94.11          & -2.02             & 94.59          & -2.11            & 82.40          & -3.92           & 80.74          & -3.88            \\
FedNH          & 94.68          & -1.75             & 95.64          & -1.45            & 83.81          & -3.71           & 81.29          & -4.59            \\
Proposed       & \textbf{95.12} & -1.78             & \textbf{95.83} & -1.66            & \textbf{85.61} & -3.66           & \textbf{83.27} & -4.51            \\
\hline\hline
\end{tabular}
\end{table}

In this experiment, we evaluate the performance of FL methods when applied to a previously unseen community. Building on the main experiments, we conduct a cross-community evaluation using the same dataset and the original train-test splits. In each fold, four communities participate in federated training using their respective training sets, while the remaining community is fully held out during training.
After federated training, each method is allowed to perform a short local adaptation using 5\% of the labeled prosumer-days from the training set of the held-out community. The test set of the held-out community remains unchanged and is used for evaluation. We compare FedAvg, BalanceFL, FedNH, and the proposed framework based on the same local detection model.

The performance results are summarized in Table~\ref{tab:cross_com}, where each $\Delta$ value denotes the relative change with respect to the corresponding result in the main experiment (Table~\ref{tab:table2}). Overall, all federated methods experience a performance decrease when evaluated on the held-out community, which is expected since the target community does not participate in federated training. Among all compared approaches, the proposed framework consistently achieves the best performance across all evaluation metrics. Moreover, its performance degradation remains moderate across metrics compared with the other federated baselines.
These results indicate that the shared components learned during federated training can provide a more effective initialization for adapting to new communities under limited local data availability.

\subsubsection{Detection Performance under Different Fraud Rate Settings}

\begin{figure}[t]
    \centering   
    \includegraphics[width=.48\textwidth]{figures/crop_curve3_legend.pdf}
    \subfigure{
        \includegraphics[width=.22\textwidth]{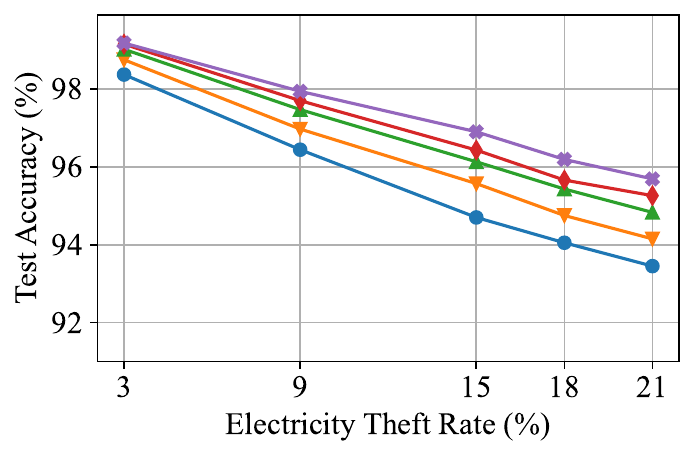}  
        \label{Fig:curve3_acc_y}
        }
    \subfigure{
        \includegraphics[width=.22\textwidth]{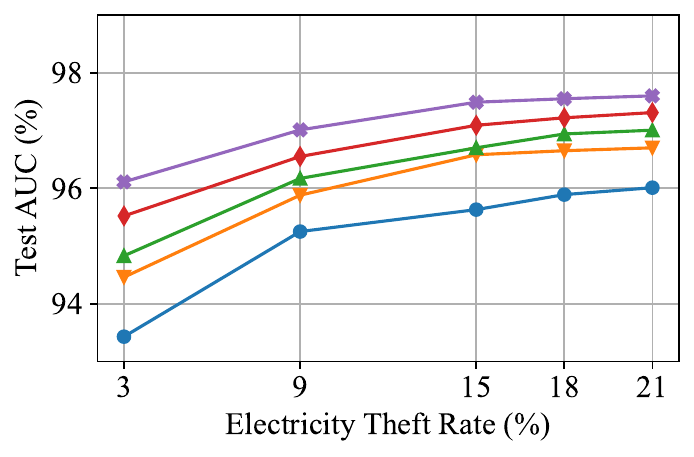} 
        \label{Fig:curve3_auc_y}
        }
    \subfigure{
        \includegraphics[width=.22\textwidth]{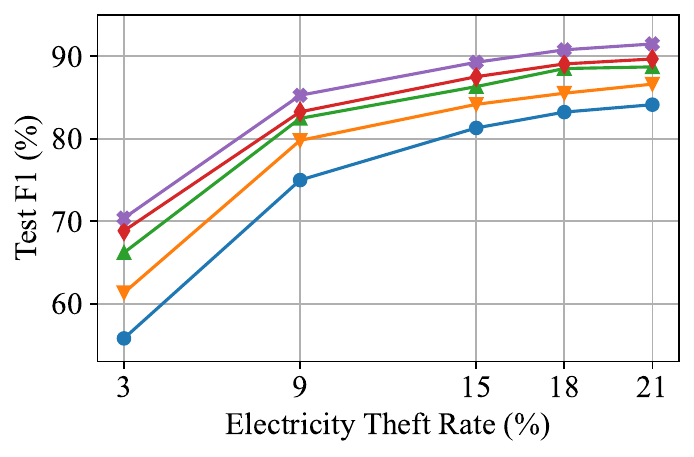}  
        \label{Fig:curve3_f1_y}
        }
    \subfigure{
        \includegraphics[width=.22\textwidth]{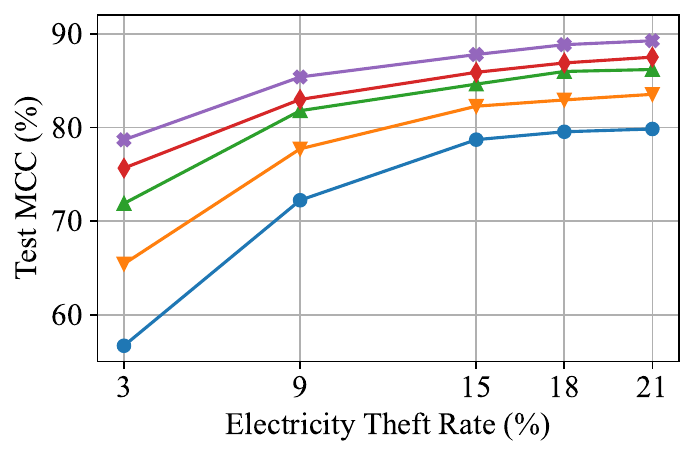}  
        \label{Fig:curve3_mcc_y}
        }
    \caption{Performance of FL frameworks under varying PV generation fraud rates.}
    \label{Fig:all_curve3}
\end{figure}

\begin{table*}[t]
\setlength{\tabcolsep}{3pt}
\renewcommand{\arraystretch}{1.2}
\centering
\caption{Performance of local models under FedAvg on the dataset with a 0.5\% fraud rate.}
\label{tab:precision_1}
\begin{tabular}{ccccccc}
\hline\hline
Local Model             & Acc. (\%)          & AUC (\%)          & F1 (\%)          & MCC (\%)          & Precision@5/1000 (\%)          & FPR@5/1000 (\%)         \\ \hline
LSTM                    & 99.23              & 82.67             & 10.73            & 10.85             & 21.83                  & 0.39                    \\
CNN-LSTM                & 99.26              & 82.41             & 13.42            & 13.67             & 24.14                  & 0.38                    \\
Transformer             & 99.47              & 86.27             & 38.66            & 40.04             & 49.43                  & \textbf{0.25}           \\
Reformer                & 99.44              & 86.15             & 35.76            & 36.90             & 47.12                  & 0.26                    \\
DLinear                 & 99.43              & 83.07             & 33.78            & 35.12             & 44.82                  & 0.28                    \\
Proposed                & \textbf{99.48}     & \textbf{87.10}    & \textbf{41.05}   & \textbf{42.42}    & \textbf{50.57}         & \textbf{0.25}           \\ \hline\hline
\end{tabular}
\end{table*}

\begin{table*}[t]
\setlength{\tabcolsep}{3pt}
\renewcommand{\arraystretch}{1.2}
\centering
\caption{Performance of FL frameworks using the proposed local model on the dataset with a 0.5\% fraud rate.}
\label{tab:precision_2}
\begin{tabular}{ccccccc}
\hline\hline
FL Framework          & Acc. (\%)          & AUC (\%)          & F1 (\%)          & MCC (\%)          & Precision@5/1000 (\%)          & FPR@5/1000 (\%) \\ \hline
Local-only            & 99.42              & 86.26             & 30.34            & 31.81             & 35.63                  & 0.32                    \\
FedAvg                & 99.48              & 87.10             & 41.05            & 42.42             & 50.57                  & 0.25                    \\
BalanceFL             & 99.55              & 87.80             & 48.32            & 50.27             & 55.17                  & 0.23                    \\
FedNH                 & 99.64              & 90.14             & 58.67            & 60.96             & 62.06                  & 0.19                    \\
Proposed              & \textbf{99.68}     & \textbf{90.25}    & \textbf{63.08}   & \textbf{65.76}    & \textbf{65.51}         & \textbf{0.17}           \\ \hline\hline
\end{tabular}
\end{table*}

Fig.~\ref{Fig:all_curve3} presents the performance of different FL frameworks under varying PV generation fraud rates. As the fraud rate increases, all methods exhibit clear and consistent performance trends across the four evaluation metrics.
In particular, F1-score and MCC increase steadily with higher fraud rates, reflecting improved detection performance as fraudulent samples become more frequent in the dataset. In contrast, accuracy shows a moderate decreasing trend, while AUC remains relatively stable across different fraud rates. This divergence indicates that accuracy and AUC are less sensitive to changes in the proportion of fraudulent samples compared to F1-score and MCC.
Across all fraud rate settings, the proposed FL framework consistently achieves the highest F1-score and MCC. Moreover, all FL frameworks outperform the Local-only baseline at each fraud rate.

In addition to the above experiments, we further consider a low-fraud-rate PVG-FD scenario in which fraudulent cases are very rare. Under such settings, conventional evaluation metrics such as accuracy and AUC tend to provide limited discrimination among methods, since most prosumer-days correspond to normal behaviors.
To better reflect detection performance in this case, we additionally report budget-aware metrics that focus on the top-ranked alerts, namely Precision@5/1000 and the corresponding false-positive rate (FPR@5/1000). These metrics evaluate how well a method performs when only a very small number of alerts can be issued.
Table~\ref{tab:precision_1} reports the results of local detection models under the FedAvg framework at a fraud rate of 0.5\%, where the proposed local model achieves the highest precision under the same alert budget with a low false-positive rate. Table~\ref{tab:precision_2} further compares different FL frameworks using the proposed local model, and the proposed framework attains the highest precision and the lowest false-positive rate among the compared approaches.
The absolute performance under the 0.5\% fraud-rate setting nevertheless remains relatively limited, which reflects the substantially increased difficulty of PVG-FD under extreme class imbalance. When the fraud rate becomes very low, the number of fraud samples is highly limited, making it more difficult for the model to learn representative fraud characteristics effectively. Meanwhile, since normal samples dominate the dataset, even a small number of false positives may significantly affect precision. This challenge becomes more evident for Precision@5/1000, since it depends not only on distinguishing fraud samples from normal samples, but also on ranking the fraud samples consistently at the very top under a limited alert budget.

\subsubsection{Case Study: Daily Fraud Probability Visualization}

\begin{figure*}[t]
\centering
\includegraphics[width=0.9\textwidth]{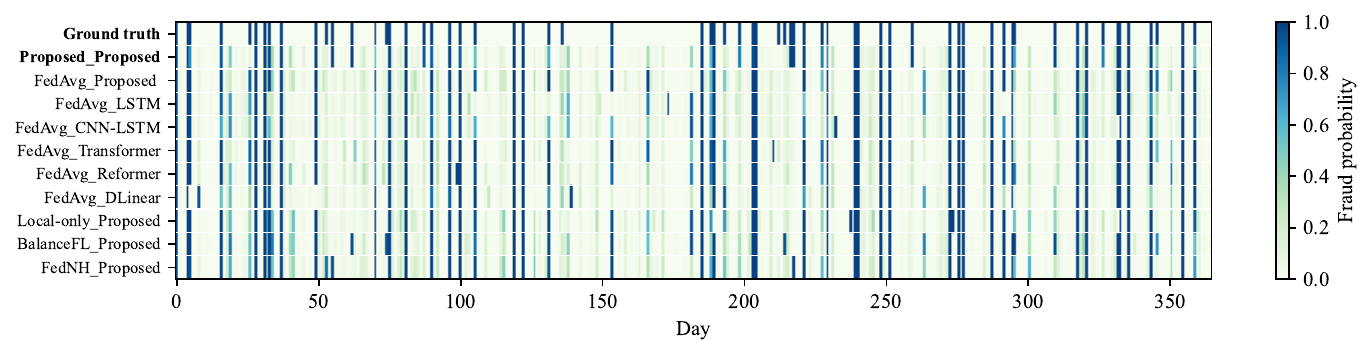}
\caption{Daily fraud probabilities for prosumer ID 194 over one year, with ground truth and predictions from different federated configurations.}
\label{fig:curve_heatmap}
\end{figure*}

To provide an intuitive illustration of detection behavior at the residence level, Fig.~\ref{fig:curve_heatmap} visualizes the daily fraud probabilities over an entire year for a prosumer. The heatmap reports the predicted fraud probability for each of the 365 days, with values ranging from 0 to 1. The top row corresponds to the ground truth, where fraud days are shown in deep blue and normal days appear nearly white.
Each subsequent row presents the predictions from one detection configuration, with the row label \texttt{FL framework\_local model} indicating the specific combination used in the main experiments. Darker colors represent higher estimated fraud probabilities, while lighter colors indicate normal behavior. This visualization enables a direct comparison of how different detection configurations respond to daily fraud events over a long time horizon and offers a qualitative view of their ability to highlight suspicious days.

\subsubsection{Discussion of Experimental Results}

The above experiments provide additional observations beyond the direct performance comparisons and help further clarify the practical significance of the reported results. From the perspective of local detection models, the standard evaluation, the cross-season evaluation, and the noisy irradiance setting together show that detection performance is affected not only by model choice, but also by seasonal variation and the quality of irradiance information. In particular, the cross-season results indicate that changes in seasonal conditions can noticeably influence detection difficulty, while the noisy irradiance results further show that degraded exogenous information can also lead to performance degradation.

From the perspective of federated framework design, the standard framework comparison, the community-number evaluation, and the cross-community setting provide further insights into factors influencing framework effectiveness in federated PVG-FD. In addition to the average performance gains over local-only training, these results show that framework performance is also influenced by how well a method handles fragmented local data and whether it can provide a useful basis for adaptation to a new community with limited labeled data.

The supplementary results under the 0.5\% fraud-rate setting further show that PVG-FD remains highly challenging when fraud cases are extremely rare. Although the proposed method still achieves the best performance among the compared approaches under this setting, the absolute performance remains relatively limited, indicating that there is still room for further improvement. These results also suggest several directions for improving detection performance under very low fraud rates. One direction is to explore learning strategies that are better suited to extreme class imbalance. Another direction is to consider optimization strategies that are more closely aligned with limited inspection-budget scenarios, since Precision@5/1000 depends not only on correct classification but also on the ability to rank fraud samples consistently at the top. In addition, incorporating richer contextual information may improve the distinction between fraud samples and normal samples in this challenging setting.

Despite the above results, the current design still has certain limitations. In the present setting, the proposed framework follows a standard synchronous FL protocol in which all participating communities exchange updates in every communication round. Therefore, the framework still requires full communication throughout training and does not yet incorporate adaptive communication-efficient strategies.

\section{Conclusion}\label{Sec:C}
In this paper, we propose a privacy-preserving distributed method for PVG-FD based on FL. Our method leverages the communication of model parameters and intermediate representations, rather than raw data, thereby safeguarding data privacy while enabling collaborative training among distributed communities. Our method consists of an FL framework and a residential-level detection model as a local model. The local model employs a co-attention mechanism to fuse PV generation and solar irradiance data, enhancing its ability to detect fraud-related discrepancies. To address class imbalance, a prototype alignment regularization term is introduced to improve the representation quality of fraud samples, mitigating the impact of imbalanced data.
Extensive experiments validate the effectiveness of the proposed method, demonstrating high detection accuracy and stable performance under class-imbalanced settings while preserving privacy.

In future work, semi-supervised and unsupervised learning approaches could be explored to reduce the reliance on labeled fraud data, improving model adaptability in data-limited or privacy-sensitive contexts. Integrating these methods with FL may also enhance PVG-FD performance.
In addition, more communication-efficient FL strategies could be explored to reduce full-round communication overhead and improve the practicality of distributed deployment.

\bibliography{reference.bib}
\bibliographystyle{IEEEtran}

\begin{IEEEbiography}[{\includegraphics[width=1in,height=1.25in,clip,keepaspectratio]{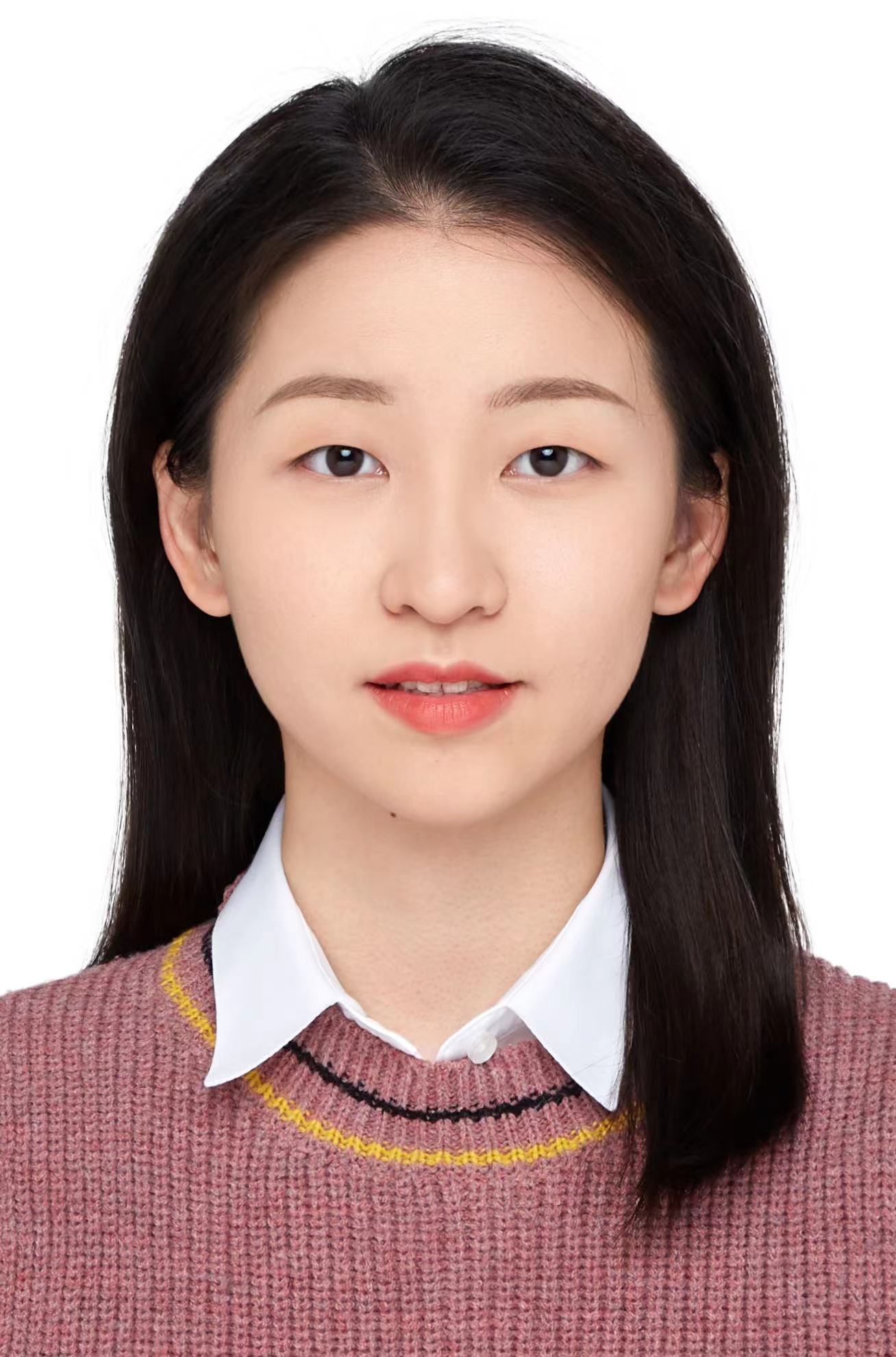}}]{Xiaolu Chen}
received the B.E. degree in Cyberspace Security and the M.S. degree in Computer Technology from the School of Computer Science and Engineering, University of Electronic Science and Technology of China. She is currently pursuing the Ph.D. degree with the Faculty of Information Technology, Monash University. Her research interests include deep learning, federated learning, and smart grids.
\end{IEEEbiography}

\begin{IEEEbiography}
[{\includegraphics[width=1in,height=1.25in,clip,keepaspectratio]{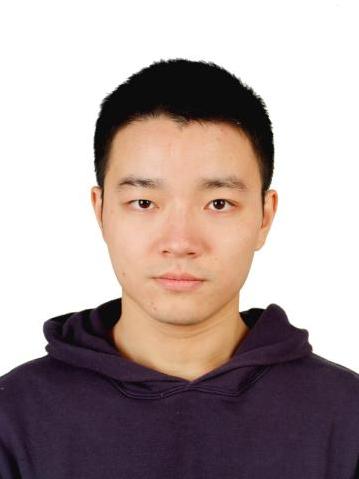}}] 
{Chenghao Huang}
received the B.E. degree in Software Engineering and the M.S. degree in Computer Science and Engineering from the University of Electronic Science and Technology of China . He is currently pursuing the Ph.D. degree with the Faculty of Information Technology, Monash University. His research interests include deep learning, federated learning, reinforcement learning and smart grids.
\end{IEEEbiography}

\begin{IEEEbiography}[{\includegraphics[width=1in,height=1.25in,clip,keepaspectratio]{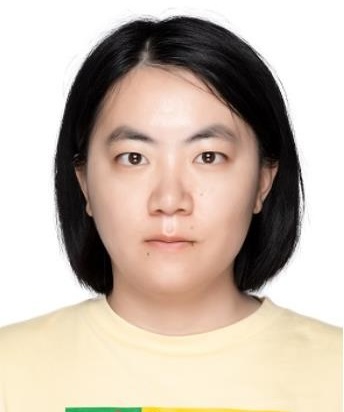}}]{Yanru Zhang}
(S’13-M’16) received the B.S. degree in electronic engineering from the University of Electronic Science and Technology of China (UESTC) in 2012, and the Ph.D. degree from the Department of Electrical and Computer Engineering, University of Houston (UH) in 2016. She worked as a Postdoctoral Fellow at UH and the Chinese University of Hong Kong successively. She is currently a Professor with the Shenzhen Institute for Advanced Study and School of Computer Science, UESTC. Her current research involves game theory, machine learning, and deep learning in network economics, Internet and applications, wireless communications, and networking. She received the Best Paper Award at IEEE SmartGridComm 2025, IEEE HPCC 2022, DependSys 2022, ICCC 2017, and ICCS 2016. 
\end{IEEEbiography}

\begin{IEEEbiography}
[{\includegraphics[width=1in,height=1.25in,clip,keepaspectratio]{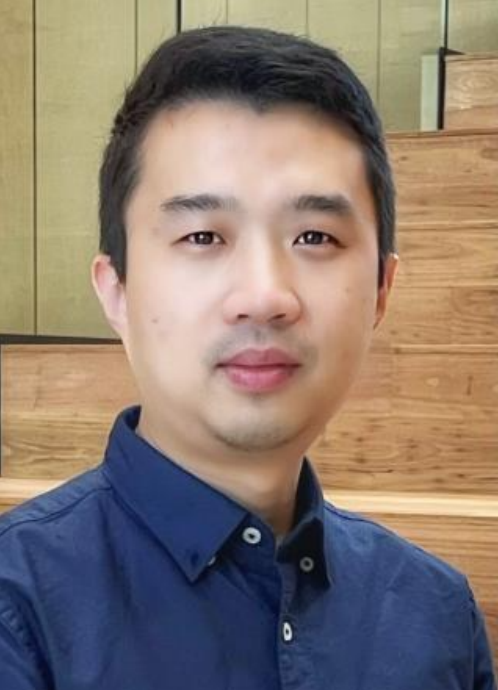}}]{Hao Wang} (M'16) received his Ph.D. in Information Engineering from The Chinese University of Hong Kong, Hong Kong, in 2016. He was a Postdoctoral Research Fellow at Stanford University, Stanford, CA, USA, and a Washington Research Foundation Innovation Fellow at the University of Washington, Seattle, WA, USA. He is currently a Senior Lecturer in the Department of Data Science and AI, Faculty of IT, Monash University, Melbourne, VIC, Australia. He received Washington Research Foundation Innovation Fellowship 2016-2018, ARC Discovery Early Career Researcher Award (DECRA) 2023-2025, First Prize Paper Award from IEEE Industry Applications Magazine 2024, Best Vision Paper Award from ACM SIGSPATIAL 2024, Best Paper Award at IEEE SmartGridComm 2025, Best Special Theme Paper at EMNLP 2025, Best Paper Award at IEEE EPEC 2025, IJPEM-GT Award from International Journal of Precision Engineering and Manufacturing-Green Technology in 2025. His research interests include optimization, machine learning, and data analytics for power and energy systems.
\end{IEEEbiography}

\vfill

\end{document}